%% file: main.tex
\documentclass[11pt,a4paper]{article}
\usepackage{times,latexsym}
\usepackage{url}
\usepackage[T1]{fontenc}

\usepackage{graphicx}      
\usepackage{booktabs}	   
\usepackage{fontawesome}   
\usepackage{color}         
\usepackage{amssymb}       
\usepackage{pgfplots}      
\usepackage{pgfplotstable} 
\usetikzlibrary{patterns}  
\usepackage{arydshln}      
\usepackage{stix}          
\usepackage{xcolor}        

\colorlet{yellow}{yellow!90!black}
\colorlet{red}{red!90}

\usepackage[acceptedWithA]{tacl2021v1}

\usepackage{xspace,mfirstuc,tabulary}

\newif\iftaclinstructions
\taclinstructionsfalse 
\iftaclinstructions
\renewcommand{\confidential}{}
\renewcommand{\anonsubtext}{(No author info supplied here, for consistency with
TACL-submission anonymization requirements)}
\newcommand{\instr}
\fi

\iftaclpubformat 

\else

\fi

\title{Language Varieties of Italy: Technology Challenges and Opportunities}

\author{
  Alan Ramponi
  \\
  Fondazione Bruno Kessler
  \\
  Trento, Italy
  \\
  \texttt{alramponi@fbk.eu}
}

\date{}

\begin{document}
\maketitle
\begin{abstract}
  Italy is characterized by a one-of-a-kind linguistic diversity landscape in Europe, which implicitly encodes local knowledge, cultural traditions, artistic expressions and history of its speakers. However, most local languages and dialects in Italy are at risk of disappearing within few generations. 
  The NLP community has recently begun to engage with endangered languages, including those of Italy. Yet, most efforts assume that these varieties are \emph{under-resourced} language monoliths with an established written form and homogeneous functions and needs, and thus highly interchangeable with each other and with \emph{high-resource}, standardized languages.
  In this paper, we introduce the linguistic context of Italy and challenge the default \emph{machine-centric} assumptions of NLP for Italy's language varieties. 
  We advocate for a shift in the paradigm from \emph{machine-centric} to \emph{speaker-centric} NLP, and provide recommendations and opportunities for work that prioritizes languages and their speakers over technological advances. To facilitate the process, we finally propose building a local community towards responsible, participatory efforts aimed at supporting vitality of languages and dialects of Italy.
\end{abstract}

\section{Introduction}

\begin{quote}
    ``\emph{Italy holds especial treasures for linguists. There is probably no other area in Europe in which such a profusion of linguistic variation is concentrated into so small a geographical area.}''
    
    ~~~~~~~~~~~~~~~~~~~~~~---~\citet{maiden1997dialects}
\end{quote}

\noindent Language is a primary means for communication that is intrinsic to the expression of culture. Through languages, we signal our social identities and convey part of our heritage~\citep{thomason-2015-endangered}. However, according to the UNESCO \emph{Atlas of World's Languages in Danger}~\citep{moseley2010atlas} about half of the spoken languages in the world are at risk of disappearing by the end of the century. Ultimately, this will lead to a loss of an integral part of cultures and traditions~\citep{hale-etal-1992-endangered}. 

The natural language processing (NLP) community has recently started to include endangered languages in its repertoire, and language varieties of Italy are no exception. However, most of the efforts in NLP implicitly assume that these language varieties are just \emph{under-resourced} entities (in terms of \emph{written} data availability) with an established written form, and with the same functions and technological needs of \emph{high-resource} standardized languages with institutional support, such as Italian or English~\citep{bird-2022-local}.
This \emph{machine-centric} approach not only fails to acknowledge that most endangered languages are primarily oral, without a standardized orthography and canonical variant, often code-switched with a co-territorial ``high-prestige'' standardized language, and serving different language functions to other languages within the local linguistic ecosystem~\citep{fishman-2001-why}, but also disregards \emph{what} and \emph{how} technologies should be built to safeguard endangered languages in the interest of speech communities~\citep{bird-2020-decolonising,caselli-etal-2021-guiding}.

In this paper, we discuss the technology challenges and opportunities for \textbf{language varieties of Italy}, one of the most linguistically-diverse landscapes in Europe which, according to UNESCO~\citep{moseley2010atlas}, currently counts over 30 languages in danger. Italy's languages and dialects are not only many for such a small area~\citep{maiden1997dialects}, but they are also very different from each other, and their linguistic distance does not typically relate to geographical distance~\citep{avolio-2009-lingue}. Most of these varieties are Romance, albeit Germanic, Slavic, Albanian and Hellenic ones also shape the Italian linguistic landscape.
As for the majority of endangered languages, most of Italy's language varieties comprise many local variants, have no standardized written form, and are just occasionally written, insofar as they are primarily used in spoken, informal settings. They typically exist in a peculiar diglossic situation with Italian, and vary in terms of recognition, protection, economic incentives, and prospects.

After introducing the linguistic situation in Italy (Section~\ref{sec:background-languages}), we review efforts in NLP for its languages and dialects (Section~\ref{sec:background-computational}). We then discuss the \emph{machine-centric} assumptions of the default NLP approach when dealing with these varieties, namely the exaggerated focus on ``machine-readable'' written data, the little regard for the representativeness of such materials of speech communities, and the homogeneous view of functions, uses, and needs across language varieties (Section~\ref{sec:challenges-directions}). We argue that language varieties of Italy should not be approached as a \emph{data commodity} for machine learning advances, and that technology should serve language varieties and their speakers and not the other way round. We thus present recommendations and opportunities for \emph{speaker-centric} NLP and advocate for a local community aimed at responsibly supporting vitality of Italy's varieties through sensitization on ethical engagement, sharing of practices, participatory collaboration, and active awareness-raising (Section~\ref{sec:opportunities}). Finally, we provide our conclusions (Section~\ref{sec:conclusion}).

\paragraph{Contributions} \emph{i)} We expose the NLP community to endangered language varieties of Italy, \emph{ii)} survey computational work for these varieties, and \emph{iii)} shed light on the main assumptions and shortcomings of the standard machine-centric NLP approach. \emph{iv)} We then identify directions and opportunities for responsible, speaker-centric efforts aimed at preserving language varieties of Italy. Finally, \emph{v)} we call for a local, multidisciplinary community that supports participatory work and knowledge sharing towards common goals. We hope our recommendations will be useful for the safeguarding of other endangered languages, too.


\section{Linguistic Context of Italy} \label{sec:background-languages}

\subsection{History and Standard Italian} \label{sec:standard-italian}

Italy is one of the most diverse landscapes in Europe in terms of language varieties~\citep{avolio-2009-lingue}. Unified late, the country was previously a collection of states with their own local languages. After the political unification in 1861, Standard Italian (ISO 639-3 code: \texttt{ita}) was adopted by the state as the official language, making it a unifying element. Italian emerged from a literary language based on Vulgar Latin, and specifically from the Tuscan variety as spoken by the Florentine upper-class society~\citep{maiden1997dialects}. At unification time, Italian was spoken by less than 10\% of the population~\citep{demauro-1963-storia}, and rates of literacy remained low for over a century, especially in rural areas. Along with education, the rise of mass media played a crucial role in establishing the widespread use of Standard Italian, mirrored by a substantial decline in the use of local languages.\footnote{Estimates indicate that 45.9\% of the population mainly speak Italian at home, 32.3\% use Italian and a local language, and 14.1\% mostly speak a local language~\citep{istat-2017-uso}.} Nowadays, Italian is the fourth most widely spoken Romance language in the world with about 68M speakers~\citep{eberhard-2022-ethnologue}.


\subsection{Languages and Dialects of Italy} \label{sec:varieties-of-italy}

Despite the establishment of Italian as national language, many local languages and dialects are still currently spoken in Italy. In Table~\ref{tab:endangered-languages} we report the language varieties of Italy classified as endangered by UNESCO~\citep{moseley2010atlas} along with their ISO 639-3 code (wherever available), linguistic branch, level of endangerment, number of speakers, and whether they have a standardized written form. 

\begin{table*}[ht!]
\resizebox{1\linewidth}{!}{%
    \centering
    \begin{tabular}{lllcr|lllcr}
        \toprule
        \textbf{Id} & \textbf{Name} & \textbf{Branch} & \textbf{LoE} & \textbf{Speakers} & \textbf{Id} & \textbf{Name} & \textbf{Branch} & \textbf{LoE} & \textbf{Speakers} \\
        \midrule
        \texttt{nap} & Neapolitan & Romance & \textcolor{yellow}{\Large$\mdlgwhtcircle$} & 6.6M &
        
        \emph{\texttt{roa-via}} & Vivaro-Alpine Occitan & Romance & \textcolor{orange}{\Large$\circledbullet$} & 65K \\

        \texttt{scn} & Sicilian & Romance & \textcolor{yellow}{\Large$\mdlgwhtcircle$} & 4.7M &
        
        \emph{\texttt{roa-gis}} & Gallo-Italic of Sicily & Romance & \textcolor{orange}{\Large$\circledbullet$} & 60K \\

        \texttt{vec} & Venetian$^{\diamondsuit}$ & Romance & \textcolor{yellow}{\Large$\mdlgwhtcircle$} & 3.9M &
        
        \texttt{lld} & Ladin$^{\diamondsuit}$ & Romance & \textcolor{orange}{\Large$\circledbullet$} & 41K \\

        \texttt{lmo} & Lombard & Romance & \textcolor{orange}{\Large$\circledbullet$} & 3.5M &
        
        \emph{\texttt{grk-gri}} & Griko & Hellenic & \textcolor{red}{\Large$\mdlgblkcircle$} & 35K \\

        \texttt{egl} & Emilian & Romance & \textcolor{orange}{\Large$\circledbullet$} & 2.0M &
        
        \emph{\texttt{roa-alc}} & Algherese Catalan & Romance & \textcolor{orange}{\Large$\circledbullet$} & 34K \\

        \texttt{pms} & Piedmontese$^{\diamondsuit}$ & Romance & \textcolor{orange}{\Large$\circledbullet$} & 1.4M &
        
        \texttt{wae} & Walser & Germanic & \textcolor{red}{\Large$\mdlgblkcircle$} & 13K \\

        \texttt{rgn} & Romagnol & Romance & \textcolor{orange}{\Large$\circledbullet$} & 1.1M &
        
        \texttt{mhn} & Mòcheno & Germanic & \textcolor{orange}{\Large$\circledbullet$} & 2K \\

        \texttt{srd} & Sardinian$^{\diamondsuit}$* & Romance & \textcolor{orange}{\Large$\circledbullet$} & 1.0M &
        
        \emph{\texttt{grk-cal}} & Calabrian Greek & Hellenic & \textcolor{red}{\Large$\mdlgblkcircle$} & 1K \\

        \texttt{fur} & Friulian$^{\diamondsuit}$ & Romance & \textcolor{orange}{\Large$\circledbullet$} & 0.6M &
        
        \emph{\texttt{roa-fae}} & Faetar & Romance & \textcolor{orange}{\Large$\circledbullet$} & 1K \\

        \texttt{lij} & Ligurian$^{\diamondsuit}$ & Romance & \textcolor{orange}{\Large$\circledbullet$} & 0.5M &
        
        \texttt{svm} & Molise Slavic & Slavic & \textcolor{red}{\Large$\mdlgblkcircle$} & 1K \\

        \emph{\texttt{gem-sty}} & South Tyrolean & Germanic & \textcolor{yellow}{\Large$\mdlgwhtcircle$} & 0.3M &
        
        \emph{\texttt{sla-res}} & Resian & Slavic & \textcolor{orange}{\Large$\circledbullet$} & <1K \\

        \texttt{aae} & Arbëreshë Albanian & Albanian & \textcolor{orange}{\Large$\circledbullet$} & 0.1M &
        
        \texttt{cim} & Cimbrian & Germanic & \textcolor{orange}{\Large$\circledbullet$} & <1K \\

        \texttt{sdn} & Gallurese & Romance & \textcolor{orange}{\Large$\circledbullet$} & 0.1M &
        
        \emph{\texttt{roa-gar}} & Gardiol & Romance & \textcolor{red}{\Large$\mdlgblkcircle$} & <1K \\

        \texttt{sdc} & Sassarese & Romance & \textcolor{orange}{\Large$\circledbullet$} & 0.1M &
        
        \texttt{itk} & Judeo-Italian & Romance & \textcolor{orange}{\Large$\circledbullet$} & <1K \\

        \texttt{frp} & Francoprovençal & Romance & \textcolor{orange}{\Large$\circledbullet$} & 71K &
        
        \emph{\texttt{gem-toi}} & Töitschu & Germanic & \textcolor{red}{\Large$\mdlgblkcircle$} & <1K \\
        
        \bottomrule
    \end{tabular}
    }%
    \caption{\label{tab:endangered-languages} 
    Endangered language varieties of Italy. Levels of endangerment (LoE) are: \textcolor{yellow}{\large$\mdlgwhtcircle$} vulnerable, \textcolor{orange}{\large$\circledbullet$} definitely endangered, and \textcolor{red}{\large$\mdlgblkcircle$} severely endangered~\citep{moseley2010atlas}. Language identifiers (Id) follow ISO 639-3 codes, wherever available; if not, we use an arbitrary designator (\emph{italicized}). The number of speakers is at a country level and is mainly taken from Glottolog and Ethnologue estimates. $^{\diamondsuit}$The language variety has a standardized written form. *Sardinian is a macro-language that includes Logudorese (\texttt{src}) and Campidanese (\texttt{sro}). Notes:~Romani (\texttt{rom}) and Corsican (\texttt{cos}) are not included due to low specificity to Italy, and for Bavarian (\texttt{bar}) and Alemannic (\texttt{gsw}) we keep the local variants that are spoken in Italy, i.e., South Tyrolean (\texttt{gem-sty}) and Walser (\texttt{wae}), respectively.}
\end{table*}

While most varieties have less than 1M speakers and are definitely or severely endangered, some are still used even by younger generations in informal settings, i.e., language varieties spoken in the south and northeast areas of the Italian peninsula~\citep{istat-2017-uso}.
Just like most languages of the world, languages and dialects of Italy are primarily used in spoken contexts, and only a fraction of them have a recently established written form. Most language varieties of Italy are Romance, insofar as they locally evolved from Vulgar Latin like Standard Italian.\footnote{Indeed, in this context the frequently used ``\emph{Italian} languages/dialects'' expression is a misnomer~\citep{avolio-2009-lingue}.} The rest include non-Latin linguistic minorities from Germanic, Albanian, Hellenic, and Slavic Indo-European branches.

\begin{figure}[ht!]
    \centering
    \includegraphics[width=1\linewidth]{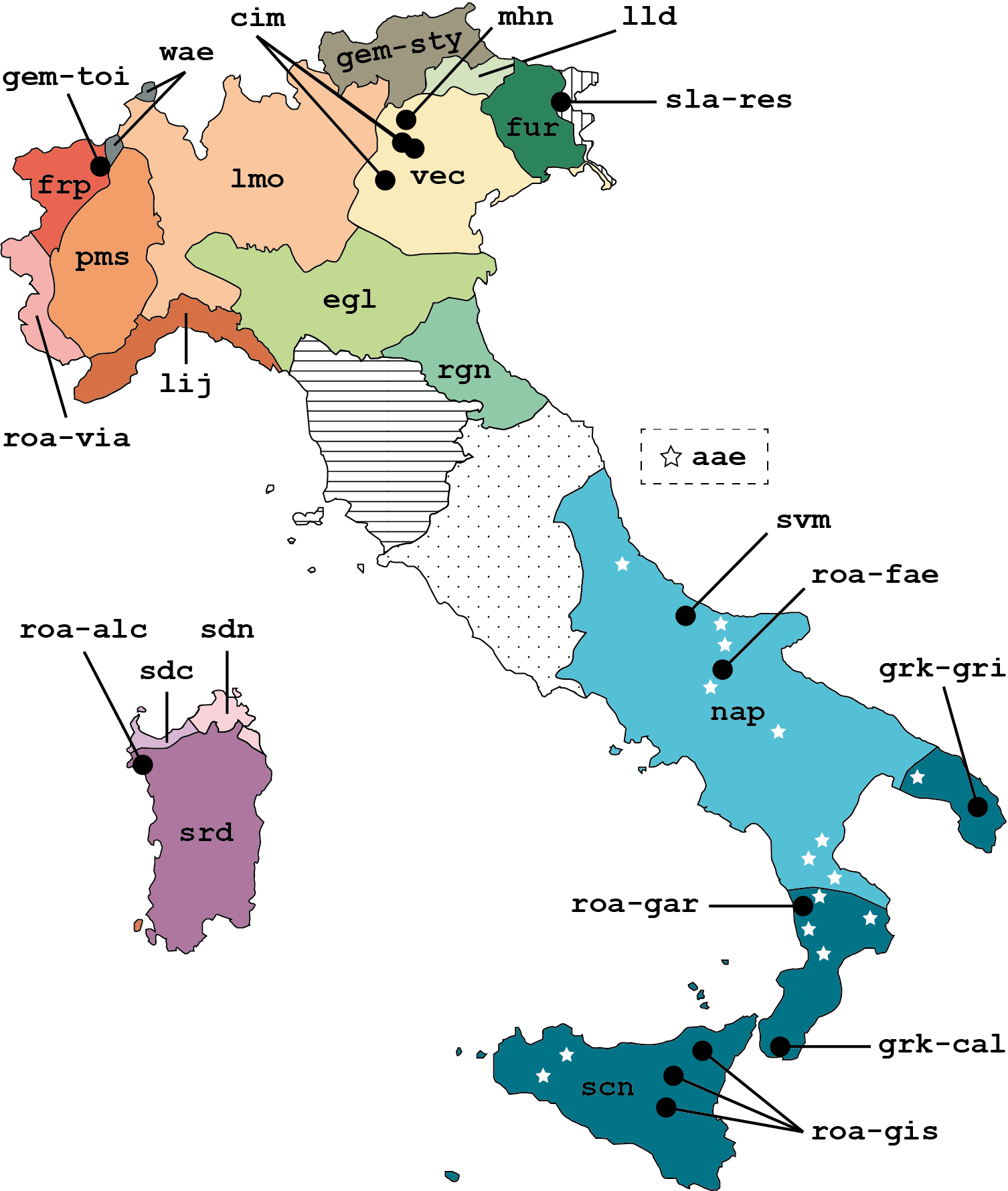}
    \caption{Map of Italy's endangered language varieties. Boundaries serve as a reading guide only: Italy's varieties often lie on a continuum without abrupt borders.}
    \label{fig:map}
\end{figure}

Due to the complex historical and sociopolitical motivations behind the use -- with negative connotation -- of the term \emph{dialetti} (``dialects'') for language varieties of Italy~\citep{avolio-2009-lingue}, and the range of meanings that the term assumes according to the context in which it is situated~\citep{berruto-2005-dialect}, we hereafter refer to those languages and dialects as \emph{language varieties}.\footnote{The term prevents any judgment on the prestige status of each variety, and avoids discussions on political matters that are not the focus of this paper.}
In the following, we contextualize endangered language varieties of Italy (Table~\ref{tab:endangered-languages}) within the linguistic macro-areas proposed in the renowned \emph{Carta dei dialetti d'Italia}~\citep{pellegrini-1977-carta}. An indicative linguistic map is also shown in Figure~\ref{fig:map}. For more details on the features of each variety and a systematic characterization of them, including local variants, we refer the reader to relevant linguistic studies and overviews on the topic~\citep[][\emph{inter alia}]{pellegrini-1977-carta,maiden1997dialects,avolio-2009-lingue}.

\paragraph{Cisalpine system} This includes Gallo-Italic varieties situated in northern Italy (i.e., \emph{Piedmontese}, \emph{Ligurian}, \emph{Lombard}, \emph{Emilian}, \emph{Romagnol}) and \emph{Venetian}, along with their many local variants.

\paragraph{Friulian system} \emph{Friulian}, a Rhaeto-Romance language recognized by the Italian state and spoken in northeast Italy, along with its local variants.

\paragraph{Tuscan system} Non-endangered language varieties that are closely related to Standard Italian (middle-northern Italy; Figure~\ref{fig:map}, \emph{horizontal lines}).

\paragraph{Middle-southern system} Non-endangered~varieties in central Italy (Figure~\ref{fig:map}, \emph{dots}), intermediate-southern varieties (e.g., \emph{Neapolitan} as a group of closely-related varieties spoken in southern continental Italy), and extreme-southern varieties (i.e., \emph{Sicilian}, its local variants, and related varieties).

\paragraph{Sardinian system} Varieties spoken in the island of Sardinia. These include the officially recognized \emph{Sardinian} macro-language (comprising \emph{Logudorese} and \emph{Campidanese}) and \emph{Gallurese} and \emph{Sassarese}, spoken in the north of the island.

\paragraph{Other varieties} These include protected varieties such as \emph{Francoprovençal} as spoken in the Aosta Valley and Piedmont and the \emph{Vivaro-Alpine Occitan} variety (all in northwest Italy), the Rhaeto-Romance \emph{Ladin} language (northeast Italy), the Austro-Bavarian \emph{South Tyrolean} variety (northern Italy), and Slovenian varieties (northeast Italy; Figure~\ref{fig:map}, \emph{vertical lines}), including \emph{Resian}. Varieties of \emph{Judeo-Italian} are also spoken across the country by very small Jewish communities.

\paragraph{Language enclaves} A number of language islands enrich the already complex linguistic landscape of Italy (Figure~\ref{fig:map}, \emph{black dots}). These include Germanic varieties in northern Italy (i.e., \emph{Cimbrian}, \emph{Mòcheno}, \emph{Walser}, \emph{Töitschu}); Modern Greek varieties in the Salento and Calabria areas, southern Italy (i.e., \emph{Griko} and \emph{Calabrian Greek}); the \emph{Molise Slavic} Serbo-Croatian variety in the Molise region, middle-southern Italy; the Francoprovençal \emph{Faetar} variety spoken in two small towns in Apulia and the Vivaro-Alpine \emph{Gardiol} enclave in the Calabria region, southern Italy; the \emph{Gallo-Italic of Sicily} Lombard enclave in the island of Sicily; the \emph{Algherese Catalan} variant spoken in Alghero (Sardinia); and \emph{Arbëreshë Albanian}, whose communities are scattered across southern Italy (Figure~\ref{fig:map}, \emph{white stars}).


\subsection{Regional Italian} \label{sec:regional-italian}

Alongside Italian and indigenous language varieties and linguistic minorities, regional varieties of Standard Italian (hereafter, \emph{regional Italian}) are also spoken by most Italian speakers. Varieties of regional Italian result from a geographical differentiation of Standard Italian after its widespread adoption, and differ from each other at various levels, i.e., syntax, morphology, phonetics, phonology, and prosody~\citep{cerruti-2011-regional,avolio-2009-lingue}. The various forms of regional Italian mostly match macro-linguistic areas of language varieties (cf.~Section~\ref{sec:varieties-of-italy}), and vary according to social and educational factors~\citep{avolio-2009-lingue}.


\section{Language Varieties of Italy and NLP}
\label{sec:background-computational}

The study, preservation, and promotion of language diversity have recently gained increasing attention in the NLP community. Initiatives such as the ACL 2022 special theme on ``\emph{Language Diversity:~from Low-Resource to Endangered Languages}''~\citep{acl-2022-association-linguistics-1}, the ACL special interest group SIGEL,\footnote{\url{https://acl-sigel.github.io/}} and relevant workshops (e.g., \textsc{ComputEL}~\citep{computel-2023-use}, \textsc{Eurali}~\citep{eurali-2022-resources}, \textsc{AmericasNLP}~\citep{americasnlp-2021-natural}) have been proposed. 
Moreover, the \textsc{VarDial} series of workshops~\citep[][\emph{inter alia}]{vardial-2023-nlp} is being routinely organized to promote the study of diatopic variation of language varieties and dialects.

In the following, we review previous work in NLP for Italy's varieties, from monolingual (Section~\ref{sec:bg-monolingual-nlp}) to multilingual efforts (Section~\ref{sec:bg-multilingual-nlp}), and highlight commonalities and differences in the shortcomings of both research lines (Section~\ref{sec:nlp-and-varieties}).


\subsection{NLP for Specific Varieties of Italy} \label{sec:bg-monolingual-nlp}

Natural language processing research for specific languages and dialects of Italy is scarce and scattered across disciplines. The most studied language variety is Venetian, for which there exist work on morphological analysis~\citep{tonelli-etal-2010-venpro2}, part-of-speech tagging~\citep[POS;][]{jaber-etal-2011-venetan}, word sense disambiguation~\citep{conforti-etal-2017-supervised} and a preliminary investigation on Venetian-English machine translation~\citep[MT;][]{delmonte-etal-2009-english}. 
Ligurian has also recently gained attention in NLP, with work on text normalization~\citep{lusito-etal-2023-text2} and the development of a Universal Dependency~\citep[UD;][]{de-marneffe-etal-2021-universal} treebank for the \emph{Genoese} variety~\citep{lusito-maillard-2021-universal}. 
A small set of \emph{Vivaro-alpine} examples has been included in an Occitan subcorpus with POS annotations~\citep{bernhard-etal-2018-corpora,bernhard-etal-2021-collecting}, whereas for Ladin, previous work includes MT from and to Italian for the \emph{Val Badia} variety~\citep{frontull-2022-machine}. 
MT has also been studied for Sicilian$\leftrightarrows$English and zero-shot Sicilian$\leftrightarrows$Italian~\citep{wdowiak-2022-recipe}, and for Italian$\rightarrow$Sardinian~\citep{tyers-etal-2017-rule} and Catalan$\rightarrow$Sardinian~\citep{fronteddu-etal-2017-eina}.

Among severely endangered varieties, Griko is the most represented in NLP. Previous work includes two Griko-Italian parallel corpora: a corpus of narratives with POS annotations~\citep{anastasopoulos-etal-2018-part2,chaudhary-etal-2021-reducing} and a small speech-derived corpus annotated with morphosyntactic, POS, glosses, and speech-related  information~\citep{boito-etal-2018-small,lekakou-etal-2013-documentation}. 
Other efforts in this space include Molise Slavic, for which field recordings, transcriptions, and Italian and German translations have been made available for the varieties of \emph{Acquaviva Collecroce}, \emph{San Felice}, and \emph{Montemitro}~\citep{breu-2017-moliseslavische}.

A number of resources have been produced for plurilingualism areas of Italy where South Tyrolean is spoken, such as a multilingual corpus of computer-mediated communication~\citep{frey-etal-2016-didi}, and a longitudinal trilingual corpus of young learners~\citep{glaznieks-etal-2022-leonide}.
Preliminary efforts such as a morphosyntactic specification for Resian~\citep{erjavec-2017-multext}, a lexical database for Sardinian, Gallurese and Sassarese~\citep{angioni-etal-2018-sardanet2}, and a tagset for Cimbrian varieties~\citep{agosti-etal-2012-curated} have also been carried out. 
There also exist a few cultural institutes that have developed tools and resources that can be interrogated online, e.g., Micurá de Rü\footnote{\url{https://www.micura.it/}} (Ladin) and Kulturinstitut Lusérn\footnote{\url{https://www.istitutocimbro.it/}} (Cimbrian), \emph{inter alia}.


\subsection{Varieties of Italy in Multilingual NLP} \label{sec:bg-multilingual-nlp}

Language varieties of Italy are increasingly represented in multilingual research.
Friulian, Ladin, Neapolitan, and Venetian have been included in the \textsc{sigmorphon} shared tasks on morphological inflection in 2018--2020~\citep{cotterell-etal-2018-conll,mccarthy-etal-2019-sigmorphon,vylomova-etal-2020-sigmorphon}, albeit the latter two have been discontinuously represented. 
More recently, a language and dialect identification shared task has been proposed~\citep{aepli-etal-2022-findings}, for which participants were given Wikipedia dumps of 11 varieties of Italy and were asked to classify text samples for a subset of the given varieties. 
Friulian, Ligurian, Lombard, Sicilian, Sardinian, and Venetian have also been included in a translation model covering 202 languages~\citep{costa-etal-2022-no}, and a corpus for cross-lingual spoken language understanding has been annotated with slot and intent information in South Tyrolean and Neapolitan~\citep{van-der-goot-etal-2021-masked,aepli-etal-2023-findings}. 

Other efforts including language varieties of Italy are sparse and mainly focus on learning methods, e.g., for learning contextualized cross-lingual word embeddings in low-resource scenarios~\citep[Griko in][]{wada-etal-2021-learning}, for language identification of text sequences in mixed-language documents~\citep[Lombard-English in][]{king-abney-2013-labeling}, or for investigating the effect of pretraining language selection on downstream zero-shot transfer~\citep[Piedmontese in][]{malkin-etal-2022-balanced}. 

Multilingual pretrained language models have also been proposed in recent times to widen language coverage in NLP, e.g., mBERT~\citep{devlin-etal-2019-bert}, mBART~\citep{liu-etal-2020-multilingual-denoising} and XLM-R~\citep{conneau-etal-2020-unsupervised}. mBERT includes some of Italy's varieties, namely Lombard, Piedmontese and Sicilian, albeit under-represented in terms of pretraining data compared to other languages. Training material is taken from entire Wikipedia editions, regardless of the covered variants, quality issues, and the representativeness of such language of speech communities (cf.~Section~\ref{sec:challenges-quality}).

\subsection{NLP Serving Varieties of Italy or the Other Way Round?} \label{sec:nlp-and-varieties}

From a closer look, we can observe that the attention to local language varieties and the very objectives of research efforts generally diverge between monolingual and multilingual NLP studies.
Most efforts for specific language varieties of Italy are explicitly intended to study or support local languages and dialects, state the orthographies and local variants being considered, and are often conducted by members of the target speech communities -- and are thus potentially driven by actual or perceived needs. 
On the other hand, recent trends in multilingual NLP are typically centered on computational advances (e.g., \emph{scaling}, \emph{generalizing}) rather than on varieties and their speakers. 
Indeed, most work in this space is driven by language technology agendas of standardized languages~\citep{bird-2022-local}, and view the \emph{under-resourcedness} of written content as a pivotal problem to be directly or indirectly fixed, do not mention which variants and orthographies of the language varieties have been included -- and why they have been chosen over the others -- perpetuating language monolithicity assumptions, and implicitly presume that language varieties are all the same in terms of functions, uses, and their speakers' needs (Section~\ref{sec:challenges-directions}).

What both research strands have in common is that the active involvement of speech communities at various stages of the design process (e.g., to express needs or assess the envisioned technology  rather than merely acting as \emph{data producers}) is typically left unspecified. This confirms similar findings by~\citet{caselli-etal-2021-guiding} and motivates us to propose new ways of working centered on language varieties and their communities (Section~\ref{sec:opportunities}).


\section{The Default \emph{Machine-centric} Approach} \label{sec:challenges-directions}

In this section, we provide an in-depth overview of the main assumptions and shortcomings of the default NLP approach -- what we refer to as \emph{machine-centric} NLP -- with a focus on language varieties of Italy. 
We first discuss the persistent focus on written data scarcity and how this is often perceived as a problem to be solved (Section~\ref{sec:challenges-data}). 
Second, we focus on widespread text collections that are typically used for training language models, arguing that the common practice of \emph{language data as a commodity} fails to represent language varieties and their speakers (Section~\ref{sec:challenges-quality}). 
Lastly, we discuss the intrinsic assumptions of the standard approach, namely the lack of regard for functions, contexts and needs of varieties (Section~\ref{sec:challenges-unstandardized}).


\subsection{Persistent Emphasis on Data Scarcity}
\label{sec:challenges-data}

A common argument in NLP work involving local language varieties of Italy is that these languages and dialects are \emph{under-resourced} in terms of ``machine-readable'' written data, and are therefore in need of more resources -- or computational means to bridge the gap -- in order to take full advantage of language technologies. 
This view not only fails to acknowledge the reasons behind written data scarcity, but also implicitly homogenizes the contexts in which such language varieties are situated and the diverse aspirations for written text -- and thus the needs for consequent technologies.

The focus on ``data quantity'' is widely rooted in the NLP community, and the amount of machine-readable language resources has also recently been used as a criterion for classifying the world's languages and highlighting their technological disparity.  
For instance, in the taxonomy of world's languages according to data availability by~\citet{joshi-etal-2020-state}, 10 endangered varieties of Italy are in the second-worst position (1: \emph{The Scraping-Bys}), while the rest belong to the worst position (0: \emph{The Left-Behinds}).\footnote{Either because explicitly indicated or not included at all.} 
Despite the best of intentions, this classification decouples the unique situation of each variety from its volume of machine-readable resources. 
While the amount of written data resources and the position in the ``technologization race'' are probably of interest to standardized and ``\emph{would-be} standardized languages''\footnote{In the context of Italy, \emph{would-be} standardized languages mostly match varieties in territories where bilingualism is officially granted by national or regional laws (Section~\ref{sec:challenges-unstandardized}).}~(see~\citet{bird-2022-local}), these factors are of little significance for most varieties, for which interests typically relate to culture preservation, language learning, and intergenerational transmission (Section~\ref{sec:opportunities}). 

Given the varied linguistic and socio-political contexts of Italy's varieties (Section~\ref{sec:challenges-unstandardized}), it is therefore more appropriate to outline \emph{which} these resources are rather than \emph{how many} they are. 
Hence, we extended the search by~\citet{joshi-etal-2020-state}, which originally included the LDC catalog,\footnote{\url{https://catalog.ldc.upenn.edu/}} ELRA Map,\footnote{\url{https://catalog.elra.info/}} and Wikipedia, by covering additional repositories and all Italy's varieties. 
We searched for main and alternate names of each variety (e.g., \texttt{nap}: \emph{Neapolitan}, \emph{Neapolitan-Calabrese}, \emph{Continental Southern Italian}) on OLAC~\citep{simons-and-bird-2003-olac}, the CLARIN Virtual Language Observatory~\citep{hinrichs-krauwer-2014-clarin}, and OPUS~\citep{tiedemann-2012-parallel}. The latter also includes data from educational resources (e.g., Tatoeba,\footnote{\url{https://tatoeba.org/}} QED~\citep{abdelali-etal-2014-amara}) and localization data from open-source software projects (e.g., Ubuntu, Gnome, Mozilla). To further include language resources that have not been submitted to mainstream repositories, we also queried Google Scholar for publications that mention both NLP and a main or alternate name of a variety. We thoroughly inspected the top 50 results for each query and retained all entries that present or use language resources. Finally, we categorized all publicly-available, curated language resources according to their language varieties, text genres, annotation types (if any), languages of parallel data (if applicable), and dataset size (Table~\ref{tab:corpora-curated}). Moreover, we inspected Wikisource, Project Gutenberg,\footnote{\url{https://www.gutenberg.org/}} UDHR,\footnote{\url{https://unicode.org/udhr/}} and raw corpora typically used in multilingual research for the presence of Italy's varieties (Table~\ref{tab:corpora-noncurated}). 

\input{tb/corpora-ann.tex}

\input{tb/corpora-general.tex}

Curated corpora for language varieties of Italy (Table~\ref{tab:corpora-curated}) greatly vary in terms of objectives, from language documentation~\citep{boito-etal-2018-small,breu-2017-moliseslavische} to supporting multilingual information access~\citep{aepli-etal-2023-findings,costa-etal-2022-no,van-der-goot-etal-2021-masked}. 
They cover a handful of language varieties (and variants, cf.~Section~\ref{sec:bg-monolingual-nlp}), are sparse in terms of text genres and annotation types, and are generally small in size. 
But \emph{(for what) is this a problem?} 
Both scarcity and sparsity of written content are a challenge to researchers embracing a machine-centric view, who may be tempted to uniformly scale current language technologies to these varieties by creating or crowd-sourcing new written corpora with annotations for a variety of tasks, design ``data-efficient'' or zero-shot methods to bridge the data scarcity gap, or just build upon raw corpora (Table~\ref{tab:corpora-noncurated}) such as web-crawled text collections regardless of how representative the content and subsequent technologies are of language varieties and speech communities (Section~\ref{sec:challenges-quality}). 
Language technologists should here take a step back and thoughtfully reflect on why there is a lack of machine-readable written resources for Italy's varieties and whether this is relevant to the target speech communities.  
By detaching from the machine-centric view of technology and engaging with speakers of local language varieties, one can realize that most languages and dialects of Italy are primarily oral, have different aspirations for written content and text-based technologies, vary in prospects according to the linguistic and socio-political contexts in which they are embedded, and serve different functions than standardized languages (Section~\ref{sec:challenges-unstandardized}). Indeed, with the exception of a few language varieties that benefit from protection, economic incentives, or co-official status with Italian, and for which a written form is used or envisioned for official purposes (Section~\ref{sec:opportunities}), written data is likely to remain scarce.


\subsection{Little Attention to Representativeness} 
\label{sec:challenges-quality}

Another assumption of the machine-centric NLP approach is that, if there is any text collection for a given language variety, it is homogeneous, representative of the community of speakers, and free of noise and boilerplate content, and therefore can be directly used for representing that language variety in language technology. However, unlike the case of standardized languages, most text collections for endangered languages naturally include content in multiple variants, freely written following no consistent or widely-established orthography (e.g., Lombard Wikipedia~\citep{miola-2017-parola}), or comprise a large amount of wrong language and non-linguistic materials~\citep{kreutzer-etal-2022-quality}. Nevertheless, those resources are typically taken monolithically regardless of their actual content.
In this section, we take Wikipedia and multilingual web-crawled corpora as case studies of mainstream text collections which are used in current NLP regardless of their representativeness of language varieties and speech communities.

Wikipedia is by far the most widely used resource in NLP when it comes to the so-called \emph{under-resourced} languages. It currently comprises content in 320 languages (as of 2023-09-10), of which 10 are endangered varieties of Italy. It additionally includes two more Wikipedias (i.e., \texttt{eml}, \texttt{roa-tar}) with deprecated or arbitrary language codes. Despite the role of Wikipedia on preserving knowledge even in lesser-used languages, the written content for most endangered varieties has to be taken carefully with regards to the varied guidelines among projects and the potential presence of fictitious and culturally-biased content. For instance, the Lombard Wikipedia leaves users freedom with respect to orthography and local variants (provided that they indicate these on the article page)~\citep{miola-2017-parola}, whereas the written content on the Piedmontese edition of Wikipedia does not match any variety actually spoken~\citep{miola-2013-sociolinguistic}. A varied use of orthography and local variants can be observed in other Wikipedia editions for Italy's varieties, such as the Ligurian Wikipedia~\citep{lusito-maillard-2021-universal}. More broadly, the content of small Wikipedias typically comprises translations of pages from larger editions (e.g., English) rather than including original content tied to speakers' identity~\citep{gobbo-and-miola-2016-modificare}. Besides objectivity, this has the effect to homogenize cultures and perspectives~\citep{callahan-and-herring-2011-cultural}. 

Near-duplicate articles are also common in Wikipedia editions for Italy's varieties. For instance, the Venetian edition of Wikipedia ($\sim$69K pages) contains placeholder content for years from 1 BC to 999 BC (1K pages) and for most of the days of the year, as well as template articles for many municipalities and provinces around the world. This suggests that a relevant portion of the encyclopedia  could be generated by bots, and thus that Wikipedia texts for Italy's language varieties not only reflect a rather artificial use of language -- what we tentatively call \emph{wikivariety} -- but also that the actual content is less than one might think.

Lastly, while \texttt{eml} has been deprecated more than 14 years ago\footnote{\url{https://iso639-3.sil.org/request/2008-040}} in favour of \texttt{egl} and \texttt{rgn} as separate ethnolinguistic entities~\citep{maiden1997dialects}, it is still in use on Wikipedia. Most \texttt{eml} pages indicate the specific variety at the top of the article, but this is rarely considered in NLP, where whole Wikipedia editions are taken as monolithic entities for training language models.

The presence of Italy's varieties on Wikipedia has an impact on the creation of web-crawled datasets. It is not surprising that multilingual corpora that include those varieties are the ones that rely on fastText LangID~\citep{joulin-etal-2017-bag}, a language identification model that currently includes a handful of Italy's language varieties and whose training material is mostly taken from Wikipedia. 

Following~\citet{kreutzer-etal-2022-quality}, who have recently highlighted systematic issues with web-crawled dataset portions for ``\emph{low-resource} languages'', we manually audit the content of crawled corpora which include Italy's varieties (cf.~Table~\ref{tab:quality-audit}) and are easily accessible. The resulting datasets are CCAligned~\citep{el-kishky-etal-2020-ccaligned}, WikiMatrix~\citep{schwenk-etal-2021-wikimatrix} (parallel) and OSCAR~\citep{abadji-etal-2022-towards} (monolingual).\footnote{We do not include XLEnt~\citep{el-kishky-etal-2021-xlent} since it comprises cross-lingual named entities rather than texts.} 

For each corpus, a native speaker of each included language variety was asked to label a random sample of 50 texts (or parallel texts, in CCAligned and WikiMatrix) according to the labeling scheme and guidelines presented in~\citet{kreutzer-etal-2022-quality}.\footnote{In some cases this was not even necessary (e.g., \texttt{lmo} in OSCAR, with just 2 instances with unambiguous sources -- namely, the \texttt{lmo} and \texttt{eml} editions of Wikipedia).} Possible labels are \textsc{c$_{nat}$} (correct, natural), \textsc{c$_{sho}$} (correct, short), \textsc{c$_{boi}$} (correct, boilerplate), \textsc{w$_{tra}$} (wrong translation -- applicable to parallel corpora only), \textsc{w$_{lan}$} (wrong language), and \textsc{w$_{nlg}$} (wrong, not language). For \texttt{lmo} and \texttt{scn} in OSCAR, the total instances are less than 50, and thus all of them have been audited. Compared to~\citet{kreutzer-etal-2022-quality}, we audit data from the latest OSCAR version (22.01), whereas for CCAligned and WikiMatrix we contribute to new language pairs (i.e., en-\texttt{srd} and it-\texttt{scn}, respectively) and report their results for en-\texttt{lmo} on WikiMatrix since we use the same data release and labeling scheme. To go beyond the approach of viewing language varieties as monoliths, native speakers were also asked to mark instances whose variants are hard to categorize because they exhibit traits of continuity with multiple varieties.\footnote{Annotations are available in our repository: \url{https://github.com/varietiesoftheboot/}}

We present the results of the audit in Table~\ref{tab:quality-audit}. For each corpus and language variety, we also report the number of texts and the percentage of samples we audited. Moreover, we indicate the percentage of Wikipedia content on audited subsets. 
OSCAR is the corpus with the highest ratio of ``correct''\footnote{As in~\citet{kreutzer-etal-2022-quality}, ``correct'' indicates that the written variant in the sample is clearly part of a language variety. It does not aim to determine a ``correct form of writing''.} content (from 50\% to 100\%); however, very few instances are included in most subsets (e.g., 2 for \texttt{lmo} and \texttt{scn}), and thus results have to be taken with a grain of salt. On the contrary, the previous OSCAR version had more instances, including additional language varieties~\citep{abadji-etal-2022-towards}, but the actual linguistic content for most of those was dramatically low (e.g., 0.0\% in-language samples for \texttt{nap}~\citep{kreutzer-etal-2022-quality}). Interestingly, the sample marked as ``wrong language'' in the \texttt{lmo} subset comes from the \texttt{eml} Wikipedia edition where it is labeled as \emph{Piacentino}, a variant of \texttt{egl} which exhibits traits of continuity with \texttt{lmo}. 
This suggests that discretizing variants into bounded languages is rather limiting since they lie on a continuum.

\input{tb/quality-audit.tex}

Regarding parallel corpora, most of the content for \texttt{srd} on CCAligned is in another language (30\%), a wrong translation (28\%), or do not even contain linguistic content (10\%). Among the 32\% ``correct'' samples, just an instance (2\%) has a clean content. The remaining 30\% contain website headers, footers and other boilerplate content. The situation is even worse on WikiMatrix: while parallel texts are cleaner than CCAligned, most pairs are not translations of each other (81.4\% on \texttt{lmo}, 78.0\% on \texttt{scn}). The ratio of ``correct'' content is thus quite low, ranging from 12.8\% to 16.0\%. 

Overall, aside from the domain-specific WikiMatrix, we observe that most of the in-language material for Italy's language varieties comes from Wikipedia articles. This suggests that content that is not already included in other resources is rarely captured, both because language identifiers trained on Wikipedia are likely to leave nothing but \emph{wikivarieties}, and most importantly because Italy's varieties are rarely written down, and if so, they are mostly code-switched with a co-territorial ``high-prestige'' standardized language with vehicular functions, e.g., Italian (Section~\ref{sec:challenges-unstandardized}).

To conclude, we argue that viewing Italy's varieties as a \emph{data commodity} for machine learning purposes without asking whether the linguistic content is representative of speech communities disregards the nature of language varieties and ignores their speakers.
We encourage researchers \emph{to care} about the varieties they work with and responsibly engage with speech communities (Section~\ref{sec:opportunities}).


\subsection{Uniform Functions, Contexts and Needs}
\label{sec:challenges-unstandardized}

The strongest assumption of the machine-centric approach is arguably to consider the diverse functions, contexts, and needs of language varieties as homogeneous -- and typically in the image of \emph{high-resource} standardized languages, e.g., Italian or English. This practice has the effect to reduce language varieties to mere linguistic codes that are dissociated from their distinctive situations.

By looking at the contemporary sociolinguistic context of Italy, most local language varieties exist in a situation of \emph{dilalìa}~\citep{berruto1987lingua} with the national language. While Italian serves as the ``high-prestige'' vehicular language~\citep{fishman-2001-why}, and it is therefore the language used in all formal settings (i.e., from education to administration), Italy's languages and dialects are primarily confined to spoken, informal situations (e.g., family, local participation), and Italian functionally overlaps with them in those informal domains -- making the situation different from the rigidly compartmentalized \emph{diglossia}~\citep{avolio-2009-lingue}. Exceptions are language varieties within territories in which bilingualism is officially granted by national laws, i.e., those of the German minority in the South Tyrol province (northern Italy), the French minority in the Aosta Valley (northwest Italy), and the Slovenian minority in some municipalities of the Friuli-Venezia Giulia region (northeast Italy). In those cases, local language varieties typically enjoy the same standing of the national language and are used (or are aimed to be used) to serve ``high-prestige'' functions. This functional differentiation should be the starting point for language technologists to reflect on the (often considered homogeneous) utility of text-based language technologies across language varieties.

The socio-political contexts in which language varieties are situated have an impact on language vitality prospects and community aspirations, too. For instance, some language varieties and their culture are protected by the~\citet{law-482-1999},\footnote{Those are the ones of the Albanian, Catalan, Germanic, Greek, Slovenian, Croatian, French, Francoprovençal, Friulian, Ladin, Occitan, and Sardinian speech communities.} albeit safeguarding measures differ on how they are locally implemented. Moreover, some of them also benefit from recognition and safeguard by regional laws,\footnote{Arbëreshë Albanian in Apulia and Calabria regions; Algherese Catalan, Gallurese, Sardinian, and Sassarese in Sardinia; German in the Walser-speaking Valle del Lys (Aosta Valley); Cimbrian, Ladin, and Mòcheno in Trentino; Calabrian Greek and Occitan (i.e., Vivaro-Alpine Gardiol) in Calabria; Francoprovençal (i.e., Faetar) and Griko in Apulia.} or are even locally co-official (i.e., German and Ladin in the South Tyrol province and French in the Aosta Valley). Finally, some language varieties are solely recognized or promoted locally, or both.\footnote{Recognized: Lombard, Piedmontese, and Sicilian in Lombardy, Piedmont, and Sicily, respectively; Promoted: Friulian and Slovenian in Friuli-Venezia Giulia, and Francoprovençal, French, Occitan, and Walser in Piedmont; Both: Venetian in Veneto and Ligurian \emph{Tabarchino} in Sardinia.} These diverse situations must be attentively considered, and engaging with local communities would allow the researcher to deeply understand how this affects their ambitions and needs (Section~\ref{sec:opportunities}).

As regards written use, although some language varieties of Italy have a notable literary tradition~\citep{avolio-2009-lingue} (e.g., works in Venetian by C.~Goldoni (18th century) and in Neapolitan by G.~Basile (16th century), \emph{inter alia}), we stress that they are nowadays primarily used in spoken, informal settings, and most of them have no standardized written form. 
Even if official orthography standards exist for some varieties, these are often unknown to speakers themselves. Indeed, in our experience speakers write ``the way words sound'' in their local variants, using just the available characters in their keyboards. Normalizing user-generated texts to a ``standard'' form~\citep[e.g.,][]{baldwin-etal-2015-shared, van-der-goot-etal-2020-norm2,van-der-goot-etal-2021-multilexnorm} has proven useful for NLP purposes, but it inevitably erases the naturally occurring sociolinguistic variation~\citep{nguyen-etal-2021-learning2}, homogenizing all variants of a language variety and imposing a ``correct'' form of writing.

But \emph{how often do speakers write in their own variety?} With the exception of restricted communities on social media and few dedicated websites, writing in some of Italy's language varieties is rather uncommon. Code-switching -- the alternation of different language varieties in a single discourse -- is instead a more widespread practice in Italy~\citep{cerruti-and-regis-2005-code}, where Standard Italian -- or any co-territorial ``high-prestige'' language in border areas -- is mixed with both Italy's varieties and regional Italian. This brings into question the utility of sentence- and document-level language identification tools supporting Italy's language varieties.


\section{Towards a \emph{Speaker-centric} Approach} \label{sec:opportunities}

The assumptions and shortcomings discussed in the preceding sections make evident that the current machine-centric approach neither respects nor represents language varieties of Italy and their speakers. Ultimately, language technology should serve speech communities and their language varieties, and not the other way round. We need to identify new ways of working that are centered on speech communities and their varieties -- what we refer to as \emph{speaker-centric} approach. 
In this section, we provide recommendations and opportunities towards speaker-centric work that foresees active engagement with speech communities. 

\paragraph{Becoming aware of local history and diverse attitudes} 
Before starting to engage with speech communities, it is advisable to become aware that local language varieties may be perceived very differently by their own speakers.
Local languages and dialects of Italy have been historically subjected to prejudices and censorship. This culminated with the \emph{Italianization} policy implemented by the fascist regime in 1923--1942 whereby ``[local language varieties] were banned in the most absolute way [...] even when playing with classmates''~\citep{camilleri2014lingua}, teaching in languages other than Italian was abolished, and foreign toponyms and surnames were changed to Italian-sounding forms. 
Among other things, this contributed over the next half century to the continued view of local language varieties as a ``synonym of ignorance and lack of integration''~\citep{dagostino2015sociolinguistica}. 
Recent years have instead witnessed an overall change in attitude on the matter, especially by the youngest, for whom local varieties are rather rediscovered as an additional expressive resource in their communicative reportoire~\citep{berruto2006quale}. 
It is therefore necessary to realize in advance that -- even within the same community -- we may encounter speakers with diverse sensitivities and motivations, and that those may also be influenced by political parties that leverage language varieties for independence purposes. 
We need to remember that speech communities do not have a ``single voice''~\citep{bird-2020-decolonising} and that language ideologies and practices may change and be embraced differently over time (e.g., Griko in~\citet{pellegrino-2021-greek}).

\paragraph{Engaging with local communities} Building relationships with speech communities~\citep{liu-etal-2022-always,schwartz-2022-primum,bird-2020-decolonising} is pivotal for speaker-centric work. It allows researchers not only to get a better sense of local communities' attitudes and aspirations, and understand the individual linguistic and socio-political contexts at the micro-level, but also to learn about local agendas to support language vitality. However, the engagement should not be for the sole benefit of the researcher, but rather based on equity, reciprocity, and respect~\citep{bird-2020-decolonising}. From here naturally comes mutual trust, deep understanding of community needs, and thus opportunities for locally-meaningful language technology applications -- that may range from online dictionaries, to computer-assisted language education, to multilingual information access, depending on the individual situation.\footnote{Indeed, it would be simplistic in the context of this paper to suggest specific language technologies for each variety.} In the context of Italy, it is important to note that the engagement process and involved actors may differ across communities. For instance, very small communities in which language varieties are mostly spoken by elders (e.g., Cimbrian, Calabrian Greek; cf.~Table~\ref{tab:endangered-languages}) are represented by a number of cultural institutes that occasionally promote initiatives on language and culture. Participating to local events, understanding customs and traditions, and ask curiosity-driven questions is probably the only way to start building meaningful bonds in this space.\footnote{To encourage researchers' awareness and participation in these contexts, we provide a collection of language and culture institutes and related entities in our repository: \url{https://github.com/varietiesoftheboot/}} 
Instead, larger speech communities of non-officially recognized varieties (e.g., Neapolitan, Venetian; cf.~Table~\ref{tab:endangered-languages}) are often supported by politically-polarized bodies, but language varieties are spoken even by younger generations~\citep{istat-2017-uso}. 
It is advisable here to engage with individuals with diverse backgrounds and demographic characteristics. 
Given the number of speakers of those varieties, if casual relationships are not already in place, bonds can be easily established in the most diverse environments, including academia. Once a collaboration space between communities and NLP researchers is found, the involvement of speech communities must not end. In the speaker-centric approach, communities are involved at multiple stages of the design process, inspired by participatory design methods~\citep{caselli-etal-2021-guiding}. External language technologists need to recall that they work with others' data for supporting vitality of others' language varieties, and that only speech communities can reliably judge the usefulness and representativeness of a given technological artifact, both during and after the process. About representativeness, it is important to acknowledge that language and culture are inseparable, and that current NLP is not culturally sensitive~\citep{hershcovich-etal-2022-challenges}. Shared knowledge may differ from place to place, and this indeed shapes language. It would not be surprising if a machine translation system for Cimbrian -- assuming that this is actually needed -- homogenized ``snow'' to a single word, regardless of the many names it gets in Cimbrian highlands according to seasons and conditions~\citep{rigonistern-1998-sentieri}. 
Broadly, this is a motivation for NLP to start shifting from the traditional, monocultural view of language to a more inclusive, culturally-aware language technology. Moreover, it opens opportunities at the intersection of participatory design and NLP, e.g., new evaluation methods based on continuous communities' feedback.

\paragraph{Building a community} 
Responsibly supporting the vitality of language varieties of Italy by adopting a speaker-centric approach could be a difficult process to initiate. Moreover, in pursing this goal we may find it valuable to build concrete relationships with other stakeholders, exchange local knowledge and experience, and establish collaborations across speech communities (e.g., those sharing similar aspirations or which language varieties are closely related) and researchers from different academic disciplines (e.g., NLP, linguistics, anthropology).
To ease this process, we initiated \textsc{Varieties of the Boot},\footnote{\url{https://varietiesoftheboot.github.io/}} a community aimed at responsibly supporting the vitality of language varieties of Italy by \emph{i)} offering guidance on the speaker-centric approach to individuals interested in engaging in this space, \emph{ii)} fostering discussion on practices that have been adopted in the past in diverse environments, lessons learnt, and mistakes to be avoided, and \emph{iii)} encouraging participatory work between diverse speech communities, cultural institutes, and fields of study. Finally, \emph{iv)} the community intends to serve as a reference point for actively raising awareness among the Italian community at large and external researchers about the often overlooked linguistic heritage of Italy. Practically, this may not only include scientific events such as thematic workshops, but also local events and communication activities on social media. 
The community opens valuable opportunities for stakeholders to learn from diverse perspectives, to responsibly engage with speech communities at different places, and to start participatory, interdisciplinary and intercultural collaborations.

\paragraph{Pursuing alternative directions} 
There are many opportunities for NLP in neighboring areas. Language technology has traditionally focused on Standard Italian, but in everyday communication Italian speakers are instead used to use their own form of regional Italian~\citep{avolio-2009-lingue} (Section~\ref{sec:regional-italian}), i.e., varieties resulting from the geographical differentiation of the standard language. 
Ultimately, NLP should better represent the actual use of the Italian language. This also opens opportunities to study fairness of current NLP models across regional variants. Moreover, NLP to study language variation and contact at scale~\citep[][\emph{inter alia}]{ramponi-casula-2023-diatopit,hovy-purschke-2018-capturing,donoso-sanchez-2017-dialectometric} can help in documenting how regional Italian varies across space. This can ultimately enrich and complement existing linguistic atlases such as ALI~\citep{pellis-massobrio-1995-atlante} and AIS~\citep{jaberg-jud-1987-ais}. Finally, based on the actual use of Italy's varieties, studying code-switching with a focus on its linguistic and social context~\citep{dogruoz-etal-2021-survey} may contribute to understanding language replacement processes~\cite{cerruti-and-regis-2005-code}.


\section{Conclusion} \label{sec:conclusion}

In this work, we present the complex linguistic landscape of Italy, shedding light on the main assumptions and shortcomings of the default, \emph{machine-centric} NLP approach for local language varieties. We advocate for a shift in the paradigm towards \emph{speaker-centric} NLP, and provide recommendations and opportunities for responsible, participatory work aimed to support vitality of language varieties of Italy, designed \emph{with} speech communities, \emph{for} serving speakers and their needs.


\section*{Acknowledgments}

We would like to thank the action editor and the anonymous reviewers for their insightful and constructive feedback during the review process. We would also like to thank Sara Tonelli and Barbara Plank for their advice on earlier versions of this paper, and the members of the Digital Humanities group at Fondazione Bruno Kessler for the precious conversations and contribution to data auditing. Further, we are grateful to Camilla Amendola for her valuable assistance in designing Figure~\ref{fig:map}.

\bibliography{anthology,tacl2021}
\bibliographystyle{acl_natbib}

\end{document}

%% file: tb/corpora-ann.tex
\begin{table*}[!ht]
    \resizebox{1\linewidth}{!}{%
    \centering
    \begin{tabular}{lllllrc}
        \toprule
        \textbf{Corpus} & \textbf{Id} & \textbf{Genre} & \textbf{Annotation} & \textbf{Parallel} & \textbf{Size} & 
        \textbf{Data} \\
        \midrule

        \textsc{xSID}~\citep{van-der-goot-etal-2021-masked} & 
        \emph{\texttt{gem-sty}} & 
        \faVolumeUp & 
        Slot; Intent & 
        \emph{multi} & 
        800 &
        \href{https://bitbucket.org/robvanderg/xsid}{URL} \\

        \textsc{UoI}~\citep{boito-etal-2018-small} & 
        \emph{\texttt{grk-gri}} & 
        \faVolumeUp & 
        Morph; POS; Glo; Sp & 
        \texttt{ita} & 
        330 &
        \href{https://github.com/antonisa/griko-italian-parallel-corpus}{URL} \\
        
        \textsc{GrikoNarrative}*~\citep{anastasopoulos-etal-2018-part2} & 
        \emph{\texttt{grk-gri}} & 
        \faBook & 
        POS & 
        \texttt{ita} & 
        942 &
        \href{https://bitbucket.org/antonis/grikoresource}{URL} \\
        
        \textsc{NormLigurian}*~\citep{lusito-etal-2023-text2} & 
        \texttt{lij} & 
        \faBook~\faMusic~\faMagic & 
        Norm & 
        \emph{std} & 
        4,394 &
        \href{https://github.com/ConseggioLigure/normalized_ligurian_corpus}{URL} \\
        
        \textsc{UD\_Ligurian-GLT}~\citep{lusito-maillard-2021-universal} & 
        \texttt{lij} & 
        \faBook~\faMagic~\faVolumeUp~\faNewspaperO~\faWikipediaW~\faCloud & 
        Morph; POS; Dep & 
        --- & 
        316 &
        \href{https://github.com/UniversalDependencies/UD_Ligurian-GLT/}{URL} \\

        \textsc{sid4lr}~\citep{aepli-etal-2023-findings} & 
        \texttt{nap} & 
        \faVolumeUp & 
        Slot; Intent & 
        \emph{multi} & 
        800 &
        \href{https://bitbucket.org/robvanderg/sid4lr}{URL} \\

        \textsc{Restaure}~\citep{bernhard-etal-2018-corpora,bernhard-etal-2021-collecting} & 
        \emph{\texttt{roa-via}} & 
        \faBook & 
        POS & 
        --- & 
        39 &
        \href{https://doi.org/10.5281/zenodo.1182948}{URL} \\
        
        \textsc{Na-našu (Acquaviva)}~\citep{breu-2017-moliseslavische} & 
        \texttt{svm} & 
        \faVolumeUp & 
        Glo; Sp & 
        \texttt{deu};\texttt{eng};\texttt{ita} & 
        890 &
        \href{https://pangloss.cnrs.fr/corpus/Na-na\%C5\%A1u_(Acquaviva_Collecroce)?lang=en}{URL} \\
        
        \textsc{Na-našu (Montemitro)}~\citep{breu-2017-moliseslavische} & 
        \texttt{svm} & 
        \faVolumeUp & 
        Glo; Sp & 
        \texttt{deu};\texttt{ita} & 
        592 &
        \href{https://pangloss.cnrs.fr/corpus/Na-na\%C5\%A1u_(Montemitro)?lang=en}{URL} \\
        
        \textsc{Na-našu (San Felice)}~\citep{breu-2017-moliseslavische} & 
        \texttt{svm} & 
        \faVolumeUp & 
        Glo; Sp & 
        \texttt{deu};\texttt{ita} & 
        628 &
        \href{https://pangloss.cnrs.fr/corpus/Na-na\%C5\%A1u_(San_Felice_del_Molise)?lang=en}{URL} \\
        
        \textsc{Stilven}~\citep{jaber-etal-2011-venetan} & 
        \texttt{vec} & 
        \faBook & 
        POS & 
        \texttt{eng} & 
        1,450 &
        \href{http://rondelmo.it/index_en.html}{URL} \\
        
        \midrule
        
        \textsc{NLLB-MD}~\citep{costa-etal-2022-no} & 
        \texttt{fur} & 
        \faNewspaperO~\faThumbsUp~\faInfoCircle & 
        --- &
        \emph{multi} & 
        8,809 &
        \href{https://github.com/facebookresearch/flores/tree/main/nllb_md}{URL} \\
        
        \textsc{NormLigurian}*~\citep{lusito-etal-2023-text2} &
        \texttt{lij} & 
        \faBook~\faWikipediaW & 
        --- &
        --- & 
        6,723 &
        \href{https://github.com/ConseggioLigure/normalized_ligurian_corpus}{URL} \\
        
        \textsc{FLORES-200}~\citep{costa-etal-2022-no} & 
        \emph{multi}$_{(6)}$ & 
        \faWikipediaW & 
        --- &
        \emph{multi} & 
        12,054 &
        \href{https://github.com/facebookresearch/flores/tree/main/flores200}{URL} \\
        
        \textsc{NLLB-Seed}~\citep{costa-etal-2022-no} & 
        \emph{multi}$_{(6)}$ & 
        \faWikipediaW & 
        --- &
        \emph{multi} & 
        37,159 &
        \href{https://github.com/facebookresearch/flores/tree/main/nllb_seed}{URL} \\
        
        \textsc{ITDI}~\citep{aepli-etal-2022-findings} & 
        \emph{multi}$_{(11)}$ & 
        \faBook~\faGlobe & 
        --- &
        --- & 
        17,886 &
        \href{https://github.com/noe-eva/ITDI_2022/}{URL} \\
        
        \textsc{Stilven}~\citep{delmonte-etal-2009-english} & 
        \texttt{vec} & 
        \faBook~\faMagic~\faNewspaperO~\faGlobe & 
        --- &
        \texttt{eng} & 
        9,027 &
        \href{http://rondelmo.it/downloads_en.html}{URL} \\
        
        \bottomrule
    \end{tabular}
    }%
    \caption{\label{tab:corpora-curated} 
    Curated corpora for language varieties of Italy used in NLP research (annotated:~\emph{top}; unannotated:~\emph{bottom}). \textbf{Corpora}. Citation and corpus name (``*'': arbitrary name). \textbf{Ids}. ISO 639-3 codes, wherever available; if not, we use an arbitrary designator (\emph{italicized}, cf.~Table~\ref{tab:endangered-languages}). \textbf{Genres}. \faBook: narratives, fiction, magazines, novels, children stories; \faMusic: poems, cantos; \faMagic: grammar examples, textbooks; \faVolumeUp: transcribed speech or field recordings; \faNewspaperO: news articles; \faWikipediaW: Wikipedia articles; \faCloud: Bible chapters; \faGlobe: quotes or proverbs from the Internet; \faThumbsUp: chat messages; \faInfoCircle: non-fiction (incl.~health reports). \textbf{Annotations}. Morph:~morphosyntactic tagging; POS:~part-of-speech tagging; Dep:~dependency parsing; Glo:~glossing; Norm:~orthographic normalization; Slot:~slot detection; Intent:~intent detection; Sp:~speech-related information (e.g., pseudo-phones, silences, etc.). \textbf{Parallels}. Language(s) of parallel data, if available (\emph{std}:~standard orthography; \emph{multi}:~many languages). \textbf{Sizes}. Number of sentences. \textbf{Data}. Link to the publicly available dataset. Notes: \emph{multi}$_{(6)}$ includes \texttt{fur}, \texttt{lij}, \texttt{lmo}, \texttt{scn}, \texttt{srd} and \texttt{vec}; \emph{multi}$_{(11)}$ additionally includes \emph{eml} (Emilian-Romagnol: \texttt{egl} and \texttt{rgn}), \texttt{lld}, \texttt{pms}, \texttt{nap} and \emph{roa-tar} (Tarantino, part of \texttt{nap}).
    }
\end{table*}

%% file: tb/corpora-general.tex
\begin{table*}[!ht]
\resizebox{1\linewidth}{!}{%
\centering
{\setlength\tabcolsep{3pt}%
\begin{tabular}{lrcccccccccccccccccccccccccccccc|ccc}
\toprule
\textbf{Corpus} & \textbf{\#Lang} & \rotatebox{90}{\texttt{aae}} & \rotatebox{90}{\texttt{cim}} & \rotatebox{90}{\texttt{egl}} & \rotatebox{90}{\texttt{frp}} & \rotatebox{90}{\texttt{fur}} & \rotatebox{90}{\texttt{gem-sty}} & \rotatebox{90}{\texttt{gem-toi}} & \rotatebox{90}{\texttt{grk-cal}} & \rotatebox{90}{\texttt{grk-gri}} & \rotatebox{90}{\texttt{itk}} & \rotatebox{90}{\texttt{lij}} & \rotatebox{90}{\texttt{lld}} & \rotatebox{90}{\texttt{lmo}} & \rotatebox{90}{\texttt{mhn}} & \rotatebox{90}{\texttt{nap}} & \rotatebox{90}{\texttt{pms}} & \rotatebox{90}{\texttt{rgn}} & \rotatebox{90}{\texttt{roa-alc}} & \rotatebox{90}{\texttt{roa-fae}} & \rotatebox{90}{\texttt{roa-gar}} & \rotatebox{90}{\texttt{roa-gis}} &  \rotatebox{90}{\texttt{roa-via}} & \rotatebox{90}{\texttt{scn}} & \rotatebox{90}{\texttt{sdc}} & \rotatebox{90}{\texttt{sdn}} & \rotatebox{90}{\texttt{sla-res}} & \rotatebox{90}{\texttt{srd}} & \rotatebox{90}{\texttt{svm}} & \rotatebox{90}{\texttt{vec}} & \rotatebox{90}{\texttt{wae}} & \rotatebox{90}{\emph{eml}} & \rotatebox{90}{\emph{roa-tar}} & \rotatebox{90}{\emph{roa-sam}} \\
\midrule

\textsc{Project Gutenberg} & 67 &  &  &  &  & \faCheck &  &  &  &  &  &  &  &  &  & \faCheck &  &  &  &  &  &  &  &  &  &  &  &  &  &  &  &  &  &   \\

\textsc{Univ.~Decl.~of Human Rights} & 529 &  &  &  & \faCheck & \faCheck &  &  &  &  &  & \faCheck & \faCheck &  &  &  &  &  &  &  &  &  &  &  &  &  &  & \faCheck &  & \faCheck &  &  &  & \faCheck \\

\textsc{Wikipedia pages} & 320 & \faHourglass[3] &  &  & \faCheck & \faCheck &  &  &  &  & \faHourglass[3] & \faCheck & \faCheck & \faCheck &  & \faCheck & \faCheck & \faHourglass[3] &  &  &  &   &   & \faCheck & \faHourglass[3] &  &  & \faCheck &  & \faCheck &  & \faCheck & \faCheck &   \\

\textsc{Wikisource pages} & 249 &  &  &  &  & \faCheck &  &  &  &  &  & \faCheck & \faCheck & \faCheck &  & \faCheck & \faCheck &  &  &  &  &  &  & \faCheck &  &  &  & \faCheck &  & \faCheck &  &  &  &   \\

\midrule

\textsc{GNOME v1} & 187 &  &  &  & \faCheck & \faCheck &  &  &  &  &  &  &  &  &  &  &  &  &  &  &  &  &  &  &  &  &  &  &  &  &  &  &  &  \\

\textsc{Mozilla-I10n v1} & 197 &  &  &  & \faCheck & \faCheck &  &  &  &  &  & \faCheck &  &  &  &  &  &  &  &  &  &  &  & \faCheck &  &  &  & \faCheck &  & \faCheck &  &  &  &  \\

\textsc{QED v2.0a} & 225 &  &  &  &  &  &  &  &  &  &  &  & \faCheck &  &  &  &  &  &  &  &  &  &  & \faCheck &  &  &  & \faCheck &  &  &  &  &  &  \\

\textsc{sardware v1} & 2 &  &  &  &  &  &  &  &  &  &  &  &  &  &  &  &  &  &  &  &  &  &  &  &  &  &  & \faCheck &  &  &  &  &  &  \\

\textsc{Tatoeba v2023-04-12} & 397 &  &  & \faCheck &  & \faCheck &  &  &  &  &  & \faCheck & \faCheck & \faCheck &  & \faCheck & \faCheck &  &  &  &  &  &  & \faCheck &  &  &  & \faCheck &  & \faCheck &  &  &  &   \\

\textsc{Ubuntu v14.10} & 244 &  &  &  & \faCheck & \faCheck &  &  &  &  &  & \faCheck & \faCheck &  &  & \faCheck & \faCheck &  &  &  &  &  &  &  &  &  &  & \faCheck &  & \faCheck & \faCheck &  &  &  \\
\midrule

\textsc{CCAligned}~\citep{el-kishky-etal-2020-ccaligned} & 137 &  &  &  &  &  &  &  &  &  &  &  &  &  &  &  &  &  &  &  &  &  &  &  &  &  &  & \faCheck &  &  &  &  &  &  \\

\textsc{CCNet}~\citep{wenzek-etal-2020-ccnet} & 130 &  &  &  &  &  &  &  &  &  &  &  &  & \faCheck &  &  & \faCheck &  &  &  &  &  &  &  &  &  &  &  &  &  &  &  &  &  \\

\textsc{OSCAR 22.01}~\citep{abadji-etal-2022-towards} & 151 &  &  &  &  &  &  &  &  &  &  &  &  & \faCheck &  &  & \faCheck &  &  &  &  &  &  & \faCheck &  &  &  &  &  &  &  & \faCheck &  &  \\

\textsc{WikiMatrix}~\citep{schwenk-etal-2021-wikimatrix} & 96 &  &  &  &  &  &  &  &  &  &  &  &  & \faCheck &  &  &  &  &  &  &  &  &  & \faCheck &  &  &  &  &  &  &  &  &  &  \\

\textsc{XLEnt v1.1}~\citep{el-kishky-etal-2021-xlent} & 120 &  &  &  &  &  &  &  &  &  &  &  &  & \faCheck &  &  &  &  &  &  &  &  &  &  &  &  &  &  &  &  &  &  &  &  \\

\bottomrule
\end{tabular}
}%
}
\caption{\label{tab:corpora-noncurated} 
Raw corpora including at least one endangered language variety of Italy. \emph{Top}:~resources with moderate verification (e.g., open collaboration projects, digitized books, and translations of international documents); \emph{Middle}:~crowd-sourced resources (i.e., educational translations and localization files from open-source software projects); \emph{Bottom}:~web-crawled resources (i.e., corpora that originate from automated web crawling and are used in multilingual NLP research). \textbf{Corpora}. Corpus name, version, and citation, if available. \textbf{\#Langs}. Number of languages in the corpus. \textbf{Presence of varieties}. \faCheck:~present; \faHourglass[3]:~upcoming (in the Wikimedia Incubator as of 2023-09-10). 
Notes:~\emph{eml} (Emilian-Romagnol) is still in use on some sources and can include either \texttt{egl} or \texttt{rgn}; \emph{roa-tar} (Tarantino) and \emph{roa-sam} (Sammarinese) are typically considered part of \texttt{nap} and \texttt{rgn}, respectively.
}
\end{table*}

%% file: tb/quality-audit.tex
\begin{table}
    \resizebox{1\linewidth}{!}{%
    \centering
    \begin{tabular}{r|r|rrr|rr}
        \toprule
        & \textsc{cca} & \multicolumn{3}{c|}{\textsc{oscar}} & \multicolumn{2}{c}{\textsc{wikimatrix}} \\
         & \texttt{srd} & \texttt{lmo} & \texttt{pms} & \texttt{scn} & \texttt{lmo} & \texttt{scn} \\
        \midrule
        \#texts & 395 & 2 & 698 & 2 & 44K & 33K \\
        \%audit & 12.7 & 100.0 & 7.2 & 100.0 & $<$0.1 & $<$0.1 \\
        \%wiki & 18.0 & 100.0 & 96.0 & 100.0 & 100.0 & 100.0 \\
        \midrule
        \textsc{c$_{nat}$} & 2.0 & 0.0 & 100.0 & 50.0 & 11.8 & 16.0 \\
        \textsc{c$_{sho}$} & 0.0 & 0.0 & 0.0 & 0.0 & 0.0 & 0.0 \\
        \textsc{c$_{boi}$} & 30.0 & 50.0 & 0.0 & 50.0 & 1.0 & 0.0 \\
        \hdashline
        \textsc{w$_{tra}$} & 28.0 & -- & -- & -- & 81.4 & 78.0 \\
        \textsc{w$_{lan}$} & 30.0 & 50.0 & 0.0 & 0.0 & 4.9 & 6.0 \\
        \textsc{w$_{nlg}$} & 10.0 & 0.0 & 0.0 & 0.0 & 1.0 & 0.0 \\
        \midrule
        \textbf{\textsc{c$_{tot}$}} & \textbf{32.0} & \textbf{50.0} & \textbf{100.0} & \textbf{100.0} & \textbf{12.8} & \textbf{16.0} \\
        \bottomrule
    \end{tabular}
    }%
    \caption{\label{tab:quality-audit} Results (\%) of the audit for Italy's language varieties on web-crawled multilingual corpora. 
    The ratio of ``correct'' samples (\textsc{c$_{tot}$}) is boldfaced.
    }
\end{table}

%% file: main.bbl
\begin{thebibliography}{93}
\expandafter\ifx\csname natexlab\endcsname\relax\def\natexlab#1{#1}\fi

\bibitem[{Abadji et~al.(2022)Abadji, Ortiz~Suarez, Romary, and Sagot}]{abadji-etal-2022-towards}
Julien Abadji, Pedro Ortiz~Suarez, Laurent Romary, and Beno{\^\i}t Sagot. 2022.
\newblock \href {https://aclanthology.org/2022.lrec-1.463} {Towards a cleaner document-oriented multilingual crawled corpus}.
\newblock In \emph{Proceedings of the Thirteenth Language Resources and Evaluation Conference}, pages 4344--4355, Marseille, France. European Language Resources Association.

\bibitem[{Abdelali et~al.(2014)Abdelali, Guzman, Sajjad, and Vogel}]{abdelali-etal-2014-amara}
Ahmed Abdelali, Francisco Guzman, Hassan Sajjad, and Stephan Vogel. 2014.
\newblock \href {http://www.lrec-conf.org/proceedings/lrec2014/pdf/877_Paper.pdf} {The {AMARA} corpus: Building parallel language resources for the educational domain}.
\newblock In \emph{Proceedings of the Ninth International Conference on Language Resources and Evaluation ({LREC}'14)}, pages 1856--1862, Reykjavik, Iceland. European Language Resources Association (ELRA).

\bibitem[{Aepli et~al.(2022)Aepli, Anastasopoulos, Chifu, Domingues, Faisal, Gaman, Ionescu, and Scherrer}]{aepli-etal-2022-findings}
No{\"e}mi Aepli, Antonios Anastasopoulos, Adrian-Gabriel Chifu, William Domingues, Fahim Faisal, Mihaela Gaman, Radu~Tudor Ionescu, and Yves Scherrer. 2022.
\newblock \href {https://aclanthology.org/2022.vardial-1.1} {Findings of the {V}ar{D}ial evaluation campaign 2022}.
\newblock In \emph{Proceedings of the Ninth Workshop on NLP for Similar Languages, Varieties and Dialects}, pages 1--13, Gyeongju, Republic of Korea. Association for Computational Linguistics.

\bibitem[{Aepli et~al.(2023)Aepli, {\c{C}}{\"o}ltekin, Van Der~Goot, Jauhiainen, Kazzaz, Ljube{\v{s}}i{\'c}, North, Plank, Scherrer, and Zampieri}]{aepli-etal-2023-findings}
No{\"e}mi Aepli, {\c{C}}a{\u{g}}r{\i} {\c{C}}{\"o}ltekin, Rob Van Der~Goot, Tommi Jauhiainen, Mourhaf Kazzaz, Nikola Ljube{\v{s}}i{\'c}, Kai North, Barbara Plank, Yves Scherrer, and Marcos Zampieri. 2023.
\newblock \href {https://aclanthology.org/2023.vardial-1.25} {Findings of the {V}ar{D}ial evaluation campaign 2023}.
\newblock In \emph{Tenth Workshop on NLP for Similar Languages, Varieties and Dialects (VarDial 2023)}, pages 251--261, Dubrovnik, Croatia. Association for Computational Linguistics.

\bibitem[{Agosti et~al.(2012)Agosti, Alber, Di~Nunzio, Dussin, Rabanus, and Tomaselli}]{agosti-etal-2012-curated}
Maristella Agosti, Birgit Alber, Giorgio~Maria Di~Nunzio, Marco Dussin, Stefan Rabanus, and Alessandra Tomaselli. 2012.
\newblock \href {http://www.lrec-conf.org/proceedings/lrec2012/pdf/290_Paper.pdf} {A curated database for linguistic research: The test case of {C}imbrian varieties}.
\newblock In \emph{Proceedings of the Eighth International Conference on Language Resources and Evaluation ({LREC}'12)}, pages 2230--2236, Istanbul, Turkey. European Language Resources Association (ELRA).

\bibitem[{Anastasopoulos et~al.(2018)Anastasopoulos, Lekakou, Quer, Zimianiti, DeBenedetto, and Chiang}]{anastasopoulos-etal-2018-part2}
Antonios Anastasopoulos, Marika Lekakou, Josep Quer, Eleni Zimianiti, Justin DeBenedetto, and David Chiang. 2018.
\newblock \href {https://aclanthology.org/C18-1214} {Part-of-speech tagging on an endangered language: {A} parallel {G}riko-{I}talian resource}.
\newblock In \emph{Proceedings of the 27th International Conference on Computational Linguistics}, pages 2529--2539, Santa Fe, New Mexico, USA. Association for Computational Linguistics.

\bibitem[{Angioni et~al.(2018)Angioni, Tuveri, Virdis, Lai, and Maltesi}]{angioni-etal-2018-sardanet2}
Manuela Angioni, Franco Tuveri, Maurizio Virdis, Laura~Lucia Lai, and Micol~Elisa Maltesi. 2018.
\newblock \href {https://aclanthology.org/2018.gwc-1.53} {{S}arda{N}et: {A} linguistic resource for {S}ardinian language}.
\newblock In \emph{Proceedings of the 9th Global Wordnet Conference}, pages 412--419, Nanyang Technological University (NTU), Singapore. Global Wordnet Association.

\bibitem[{Avolio(2009)}]{avolio-2009-lingue}
Francesco Avolio. 2009.
\newblock \emph{{Lingue e Dialetti d'Italia}}.
\newblock {Le Bussole}. {Carocci}, {Roma, Italy}.

\bibitem[{Baldwin et~al.(2015)Baldwin, de~Marneffe, Han, Kim, Ritter, and Xu}]{baldwin-etal-2015-shared}
Timothy Baldwin, Marie~Catherine de~Marneffe, Bo~Han, Young-Bum Kim, Alan Ritter, and Wei Xu. 2015.
\newblock \href {https://doi.org/10.18653/v1/W15-4319} {Shared tasks of the 2015 workshop on noisy user-generated text: {T}witter lexical normalization and named entity recognition}.
\newblock In \emph{Proceedings of the Workshop on Noisy User-generated Text}, pages 126--135, Beijing, China. Association for Computational Linguistics.

\bibitem[{Bartoli et~al.(1995)Bartoli, Pellis, and Massobrio}]{pellis-massobrio-1995-atlante}
Matteo Bartoli, Ugo Pellis, and Lorenzo Massobrio. 1995.
\newblock \emph{Atlante Linguistico Italiano}.
\newblock Istituto Poligrafico e Zecca dello Stato, Roma, Italy.

\bibitem[{Bernhard et~al.(2021)Bernhard, Ligozat, Bras, Martin, Vergez-Couret, Erhart, Sibille, Todirascu, Boula~de Mare{\"u}il, and Huck}]{bernhard-etal-2021-collecting}
Delphine Bernhard, Anne-Laure Ligozat, Myriam Bras, Fanny Martin, Marianne Vergez-Couret, Pascale Erhart, Jean Sibille, Amalia Todirascu, Philippe Boula~de Mare{\"u}il, and Dominique Huck. 2021.
\newblock \href {http://hdl.handle.net/10125/74645} {{Collecting and annotating corpora for three under-resourced languages of France: Methodological issues}}.
\newblock \emph{{Language Documentation \& Conservation}}, 15:316--357.

\bibitem[{Bernhard et~al.(2018)Bernhard, Ligozat, Martin, Bras, Magistry, Vergez-Couret, Steibl{\'e}, Erhart, Hathout, Huck, Rey, Reyn{\'e}s, Rosset, Sibille, and Lavergne}]{bernhard-etal-2018-corpora}
Delphine Bernhard, Anne-Laure Ligozat, Fanny Martin, Myriam Bras, Pierre Magistry, Marianne Vergez-Couret, Lucie Steibl{\'e}, Pascale Erhart, Nabil Hathout, Dominique Huck, Christophe Rey, Philippe Reyn{\'e}s, Sophie Rosset, Jean Sibille, and Thomas Lavergne. 2018.
\newblock \href {https://aclanthology.org/L18-1619} {Corpora with part-of-speech annotations for three regional languages of {F}rance: {A}lsatian, {O}ccitan and {P}icard}.
\newblock In \emph{Proceedings of the Eleventh International Conference on Language Resources and Evaluation ({LREC} 2018)}, Miyazaki, Japan. European Language Resources Association (ELRA).

\bibitem[{Berruto(1987)}]{berruto1987lingua}
Gaetano Berruto. 1987.
\newblock Lingua, dialetto, diglossia, dilalìa.
\newblock In \emph{Romania et Slavia Adriatica}, pages 57--81. Buske, Hamburg, Germany.

\bibitem[{Berruto(2005)}]{berruto-2005-dialect}
Gaetano Berruto. 2005.
\newblock \href {https://doi.org/10.1017/CBO9780511486623.005} {Dialect/standard convergence, mixing, and models of language contact: The case of {I}taly}.
\newblock \emph{Dialect Change: Convergence and Divergence in European Languages}, pages 81--95.

\bibitem[{Berruto(2006)}]{berruto2006quale}
Gaetano Berruto. 2006.
\newblock Quale dialetto per l’{I}talia del duemila? {A}spetti dell’italianizzazione e risorgenze dialettali in {P}iemonte (e altrove).
\newblock In \emph{Lingua e Dialetto nell’Italia del Duemila}, pages 101--127. Congedo, Galatina, Italy.

\bibitem[{Bird(2020)}]{bird-2020-decolonising}
Steven Bird. 2020.
\newblock \href {https://doi.org/10.18653/v1/2020.coling-main.313} {Decolonising speech and language technology}.
\newblock In \emph{Proceedings of the 28th International Conference on Computational Linguistics}, pages 3504--3519, Barcelona, Spain (Online). International Committee on Computational Linguistics.

\bibitem[{Bird(2022)}]{bird-2022-local}
Steven Bird. 2022.
\newblock \href {https://doi.org/10.18653/v1/2022.acl-long.539} {Local languages, third spaces, and other high-resource scenarios}.
\newblock In \emph{Proceedings of the 60th Annual Meeting of the Association for Computational Linguistics (Volume 1: Long Papers)}, pages 7817--7829, Dublin, Ireland. Association for Computational Linguistics.

\bibitem[{Boito et~al.(2018)Boito, Anastasopoulos, Villavicencio, Besacier, and Lekakou}]{boito-etal-2018-small}
Marcely~Zanon Boito, Antonios Anastasopoulos, Aline Villavicencio, Laurent Besacier, and Marika Lekakou. 2018.
\newblock \href {https://doi.org/10.21437/SLTU.2018-8} {A small {G}riko-{I}talian speech translation corpus}.
\newblock In \emph{Proceedings of the 6th International Workshop on Spoken Language Technologies for Under-Resourced Languages}, pages 36--41.

\bibitem[{Breu(2017)}]{breu-2017-moliseslavische}
Walter Breu. 2017.
\newblock \href {https://doi.org/10.2307/j.ctv11sn5zw} {\emph{Slavische Mikrosprachen im Absoluten Sprachkontakt: Glossierte und Interpretierte Sprachaufnahmen aus Italien, Deutschland, Österreich und Griechenland. Teil I: Moliseslavische Texte aus Acquaviva Collecroce, Montemitro und San Felice del Molise}}.
\newblock Harrassowitz Verlag, Wiesbaden, Germany.

\bibitem[{Callahan and Herring(2011)}]{callahan-and-herring-2011-cultural}
Ewa~S. Callahan and Susan~C. Herring. 2011.
\newblock \href {https://doi.org/10.1002/asi.21577} {Cultural bias in {W}ikipedia content on famous persons}.
\newblock \emph{Journal of the American Society for Information Science and Technology}, 62(10):1899--1915.

\bibitem[{Camilleri and De~Mauro(2014)}]{camilleri2014lingua}
Andrea Camilleri and Tullio De~Mauro. 2014.
\newblock \emph{La Lingua Batte dove il Dente Duole}, 3rd edition.
\newblock I Robinson. Letture. Laterza, Roma-Bari, Italy.

\bibitem[{Caselli et~al.(2021)Caselli, Cibin, Conforti, Encinas, and Teli}]{caselli-etal-2021-guiding}
Tommaso Caselli, Roberto Cibin, Costanza Conforti, Enrique Encinas, and Maurizio Teli. 2021.
\newblock \href {https://doi.org/10.18653/v1/2021.nlp4posimpact-1.4} {Guiding principles for participatory design-inspired natural language processing}.
\newblock In \emph{Proceedings of the 1st Workshop on NLP for Positive Impact}, pages 27--35, Online. Association for Computational Linguistics.

\bibitem[{Cerruti(2011)}]{cerruti-2011-regional}
Massimo Cerruti. 2011.
\newblock \href {https://doi.org/10.1515/ijsl.2011.028} {Regional varieties of {I}talian in the linguistic repertoire}.
\newblock \emph{International Journal of the Sociology of Language}, 2011(210):9--28.

\bibitem[{Cerruti and Regis(2005)}]{cerruti-and-regis-2005-code}
Massimo Cerruti and Riccardo Regis. 2005.
\newblock \href {https://www.italian-journal-linguistics.com/app/uploads/2021/06/08.Cerruti-Regis_01.De_.pdf} {`{C}ode switching' e teoria linguistica: La situazione italo-romanza}.
\newblock \emph{Italian Journal of Linguistics}, 17(1):179.

\bibitem[{Chaudhary et~al.(2021)Chaudhary, Anastasopoulos, Sheikh, and Neubig}]{chaudhary-etal-2021-reducing}
Aditi Chaudhary, Antonios Anastasopoulos, Zaid Sheikh, and Graham Neubig. 2021.
\newblock \href {https://doi.org/10.1162/tacl_a_00350} {Reducing confusion in active learning for part-of-speech tagging}.
\newblock \emph{Transactions of the Association for Computational Linguistics}, 9:1--16.

\bibitem[{Conforti and Fraser(2017)}]{conforti-etal-2017-supervised}
Costanza Conforti and Alexander Fraser. 2017.
\newblock \href {https://www.cis.uni-muenchen.de/~fraser/pubs/conforti_flairs2017.pdf} {Supervised word sense disambiguation for {V}enetan: A proof-of-concept experiment}.
\newblock In \emph{The Thirtieth International Flairs Conference}, Marco Island, Florida. Association for the Advancement of Artificial Intelligence.

\bibitem[{Conneau et~al.(2020)Conneau, Khandelwal, Goyal, Chaudhary, Wenzek, Guzm{\'a}n, Grave, Ott, Zettlemoyer, and Stoyanov}]{conneau-etal-2020-unsupervised}
Alexis Conneau, Kartikay Khandelwal, Naman Goyal, Vishrav Chaudhary, Guillaume Wenzek, Francisco Guzm{\'a}n, Edouard Grave, Myle Ott, Luke Zettlemoyer, and Veselin Stoyanov. 2020.
\newblock \href {https://doi.org/10.18653/v1/2020.acl-main.747} {Unsupervised cross-lingual representation learning at scale}.
\newblock In \emph{Proceedings of the 58th Annual Meeting of the Association for Computational Linguistics}, pages 8440--8451, Online. Association for Computational Linguistics.

\bibitem[{Cotterell et~al.(2018)Cotterell, Kirov, Sylak-Glassman, Walther, Vylomova, McCarthy, Kann, Mielke, Nicolai, Silfverberg, Yarowsky, Eisner, and Hulden}]{cotterell-etal-2018-conll}
Ryan Cotterell, Christo Kirov, John Sylak-Glassman, G{\'e}raldine Walther, Ekaterina Vylomova, Arya~D. McCarthy, Katharina Kann, Sabrina~J. Mielke, Garrett Nicolai, Miikka Silfverberg, David Yarowsky, Jason Eisner, and Mans Hulden. 2018.
\newblock \href {https://doi.org/10.18653/v1/K18-3001} {The {C}o{NLL}{--}{SIGMORPHON} 2018 shared task: Universal morphological reinflection}.
\newblock In \emph{Proceedings of the {C}o{NLL}{--}{SIGMORPHON} 2018 Shared Task: Universal Morphological Reinflection}, pages 1--27, Brussels. Association for Computational Linguistics.

\bibitem[{D'Agostino(2015)}]{dagostino2015sociolinguistica}
Mari D'Agostino. 2015.
\newblock Sociolinguistica dell’italiano contemporaneo.
\newblock In \emph{L'Italia e le sue Regioni}, volume III. Treccani, Roma, Italy.

\bibitem[{De~Mauro(1963)}]{demauro-1963-storia}
Tullio De~Mauro. 1963.
\newblock \emph{Storia Linguistica dell'Italia Unita}.
\newblock Laterza, Roma-Bari, Italy.

\bibitem[{Delmonte et~al.(2009)Delmonte, Bristot, Tonelli, and Pianta}]{delmonte-etal-2009-english}
Rodolfo Delmonte, Antonella Bristot, Sara Tonelli, and Emanuele Pianta. 2009.
\newblock \href {https://aclanthology.org/www.mt-archive.info/ISMTCL-2009-abstracts.htm} {English/{V}eneto resource poor machine translation with {STILVEN}}.
\newblock In \emph{Proceedings of ISMTCL}, pages 82--89, Besançon, France. Presses Universitaires de Franche-Comté.

\bibitem[{Devlin et~al.(2019)Devlin, Chang, Lee, and Toutanova}]{devlin-etal-2019-bert}
Jacob Devlin, Ming-Wei Chang, Kenton Lee, and Kristina Toutanova. 2019.
\newblock \href {https://doi.org/10.18653/v1/N19-1423} {{BERT}: Pre-training of deep bidirectional transformers for language understanding}.
\newblock In \emph{Proceedings of the 2019 Conference of the North {A}merican Chapter of the Association for Computational Linguistics: Human Language Technologies, Volume 1 (Long and Short Papers)}, pages 4171--4186, Minneapolis, Minnesota. Association for Computational Linguistics.

\bibitem[{Do{\u{g}}ru{\"o}z et~al.(2021)Do{\u{g}}ru{\"o}z, Sitaram, Bullock, and Toribio}]{dogruoz-etal-2021-survey}
A.~Seza Do{\u{g}}ru{\"o}z, Sunayana Sitaram, Barbara~E. Bullock, and Almeida~Jacqueline Toribio. 2021.
\newblock \href {https://doi.org/10.18653/v1/2021.acl-long.131} {A survey of code-switching: Linguistic and social perspectives for language technologies}.
\newblock In \emph{Proceedings of the 59th Annual Meeting of the Association for Computational Linguistics and the 11th International Joint Conference on Natural Language Processing (Volume 1: Long Papers)}, pages 1654--1666, Online. Association for Computational Linguistics.

\bibitem[{Donoso and S{\'a}nchez(2017)}]{donoso-sanchez-2017-dialectometric}
Gonzalo Donoso and David S{\'a}nchez. 2017.
\newblock \href {https://doi.org/10.18653/v1/W17-1202} {Dialectometric analysis of language variation in {T}witter}.
\newblock In \emph{Proceedings of the Fourth Workshop on {NLP} for Similar Languages, Varieties and Dialects ({V}ar{D}ial)}, pages 16--25, Valencia, Spain. Association for Computational Linguistics.

\bibitem[{Eberhard et~al.(2022)Eberhard, Simons, and Fennig}]{eberhard-2022-ethnologue}
David~M. Eberhard, Gary~F. Simons, and Charles~D. Fennig. 2022.
\newblock \emph{Ethnologue: Languages of the World}, twenty-fifth edition.
\newblock SIL International, Dallas, Texas.

\bibitem[{El-Kishky et~al.(2020)El-Kishky, Chaudhary, Guzm{\'a}n, and Koehn}]{el-kishky-etal-2020-ccaligned}
Ahmed El-Kishky, Vishrav Chaudhary, Francisco Guzm{\'a}n, and Philipp Koehn. 2020.
\newblock \href {https://doi.org/10.18653/v1/2020.emnlp-main.480} {{CCA}ligned: A massive collection of cross-lingual web-document pairs}.
\newblock In \emph{Proceedings of the 2020 Conference on Empirical Methods in Natural Language Processing (EMNLP)}, pages 5960--5969, Online. Association for Computational Linguistics.

\bibitem[{El-Kishky et~al.(2021)El-Kishky, Renduchintala, Cross, Guzm{\'a}n, and Koehn}]{el-kishky-etal-2021-xlent}
Ahmed El-Kishky, Adithya Renduchintala, James Cross, Francisco Guzm{\'a}n, and Philipp Koehn. 2021.
\newblock \href {https://doi.org/10.18653/v1/2021.emnlp-main.814} {{XLE}nt: Mining a large cross-lingual entity dataset with lexical-semantic-phonetic word alignment}.
\newblock In \emph{Proceedings of the 2021 Conference on Empirical Methods in Natural Language Processing}, pages 10424--10430, Online and Punta Cana, Dominican Republic. Association for Computational Linguistics.

\bibitem[{Erjavec(2017)}]{erjavec-2017-multext}
Toma{\v{z}} Erjavec. 2017.
\newblock \href {https://doi.org/10.1007/978-94-024-0881-2_17} {{MULTEXT}-{E}ast}.
\newblock In \emph{Handbook of Linguistic Annotation}, pages 441--462. Springer, Dordrecht, Netherlands.

\bibitem[{Fishman(2001)}]{fishman-2001-why}
Joshua~A. Fishman. 2001.
\newblock \href {https://doi.org/10.21832/9781853597060-003} {Why is it so hard to save a threatened language?}
\newblock In \emph{Can Threatened Languages be Saved?}, pages 1--22. Multilingual Matters, Bristol, Blue Ridge Summit.

\bibitem[{Frey et~al.(2016)Frey, Glaznieks, and Stemle}]{frey-etal-2016-didi}
Jennifer-Carmen Frey, Aivars Glaznieks, and Egon~W. Stemle. 2016.
\newblock \href {https://doi.org/10.4000/books.aaccademia.1782} {The {D}i{D}i corpus of {S}outh {T}yrolean {CMC} data: A multilingual corpus of {F}acebook texts}.
\newblock In \emph{Proceedings of the Third Italian Conference on Computational Linguistics CLiC-it 2016}, Napoli, Italy. Accademia University Press.

\bibitem[{Fronteddu et~al.(2017)Fronteddu, Alòs~i Font, and Tyers}]{fronteddu-etal-2017-eina}
Gianfranco Fronteddu, Hèctor Alòs~i Font, and Francis~M. Tyers. 2017.
\newblock \href {https://doi.org/10.21814/lm.9.2.255} {Una eina per a una llengua en procés d’estandardització: El traductor automàtic català-sard}.
\newblock \emph{Linguamática}, 9(2):3--20.

\bibitem[{Frontull(2022)}]{frontull-2022-machine}
Samuel Frontull. 2022.
\newblock {M}achine {T}ranslation for the {L}ow-resource {L}adin of the {V}al {B}adia.
\newblock Master's thesis, University of Innsbruck.

\bibitem[{Glaznieks et~al.(2022)Glaznieks, Frey, Stopfner, Zanasi, and Nicolas}]{glaznieks-etal-2022-leonide}
Aivars Glaznieks, Jennifer-Carmen Frey, Maria Stopfner, Lorenzo Zanasi, and Lionel Nicolas. 2022.
\newblock \href {https://doi.org/10.1075/ijlcr.21004.gla} {Leonide: A longitudinal trilingual corpus of young learners of {I}talian, {G}erman and {E}nglish}.
\newblock \emph{International Journal of Learner Corpus Research}, 8(1):97--120.

\bibitem[{Gobbo and Miola(2016)}]{gobbo-and-miola-2016-modificare}
Federico Gobbo and Emanuele Miola. 2016.
\newblock \href {http://hdl.handle.net/11585/649744} {Modificare l'immagine linguistica: Esperanto e {P}iemontese a confronto}.
\newblock In \emph{Représentations Sociales des Langues et Politiques Linguistiques. Déterminismes, Implications, Regards Croisés}, pages 287--304. Aracne Editrice, Roma, Italy.

\bibitem[{van~der Goot et~al.(2020)van~der Goot, Ramponi, Caselli, Cafagna, and De~Mattei}]{van-der-goot-etal-2020-norm2}
Rob van~der Goot, Alan Ramponi, Tommaso Caselli, Michele Cafagna, and Lorenzo De~Mattei. 2020.
\newblock \href {https://aclanthology.org/2020.lrec-1.769} {Norm it! {L}exical normalization for {I}talian and its downstream effects for dependency parsing}.
\newblock In \emph{Proceedings of the Twelfth Language Resources and Evaluation Conference}, pages 6272--6278, Marseille, France. European Language Resources Association.

\bibitem[{van~der Goot et~al.(2021{\natexlab{a}})van~der Goot, Ramponi, Zubiaga, Plank, Muller, San Vicente~Roncal, Ljube{\v{s}}i{\'c}, {\c{C}}etino{\u{g}}lu, Mahendra, {\c{C}}olako{\u{g}}lu, Baldwin, Caselli, and Sidorenko}]{van-der-goot-etal-2021-multilexnorm}
Rob van~der Goot, Alan Ramponi, Arkaitz Zubiaga, Barbara Plank, Benjamin Muller, I{\~n}aki San Vicente~Roncal, Nikola Ljube{\v{s}}i{\'c}, {\"O}zlem {\c{C}}etino{\u{g}}lu, Rahmad Mahendra, Talha {\c{C}}olako{\u{g}}lu, Timothy Baldwin, Tommaso Caselli, and Wladimir Sidorenko. 2021{\natexlab{a}}.
\newblock \href {https://doi.org/10.18653/v1/2021.wnut-1.55} {{M}ulti{L}ex{N}orm: A shared task on multilingual lexical normalization}.
\newblock In \emph{Proceedings of the Seventh Workshop on Noisy User-generated Text (W-NUT 2021)}, pages 493--509, Online. Association for Computational Linguistics.

\bibitem[{van~der Goot et~al.(2021{\natexlab{b}})van~der Goot, Sharaf, Imankulova, {\"U}st{\"u}n, Stepanovi{\'c}, Ramponi, Khairunnisa, Komachi, and Plank}]{van-der-goot-etal-2021-masked}
Rob van~der Goot, Ibrahim Sharaf, Aizhan Imankulova, Ahmet {\"U}st{\"u}n, Marija Stepanovi{\'c}, Alan Ramponi, Siti~Oryza Khairunnisa, Mamoru Komachi, and Barbara Plank. 2021{\natexlab{b}}.
\newblock \href {https://doi.org/10.18653/v1/2021.naacl-main.197} {From masked language modeling to translation: Non-{E}nglish auxiliary tasks improve zero-shot spoken language understanding}.
\newblock In \emph{Proceedings of the 2021 Conference of the North American Chapter of the Association for Computational Linguistics: Human Language Technologies}, pages 2479--2497, Online. Association for Computational Linguistics.

\bibitem[{Hale et~al.(1992)Hale, Krauss, Watahomigie, Yamamoto, Craig, Jeanne, and England}]{hale-etal-1992-endangered}
Ken Hale, Michael Krauss, Lucille~J. Watahomigie, Akira~Y. Yamamoto, Colette Craig, LaVerne~Masayesva Jeanne, and Nora~C. England. 1992.
\newblock \href {https://doi.org/10.2307/416368} {Endangered languages}.
\newblock \emph{Language}, 68(1):1--42.

\bibitem[{Harrigan et~al.(2023)Harrigan, Chaudhary, Rijhwani, Moeller, Arppe, Palmer, Henke, and Rosenblum}]{computel-2023-use}
Atticus Harrigan, Aditi Chaudhary, Shruti Rijhwani, Sarah Moeller, Antti Arppe, Alexis Palmer, Ryan Henke, and Daisy Rosenblum, editors. 2023.
\newblock \href {https://aclanthology.org/2023.computel-1.0} {\emph{Proceedings of the Sixth Workshop on the Use of Computational Methods in the Study of Endangered Languages}}. Association for Computational Linguistics, Remote.

\bibitem[{Hershcovich et~al.(2022)Hershcovich, Frank, Lent, de~Lhoneux, Abdou, Brandl, Bugliarello, Cabello~Piqueras, Chalkidis, Cui, Fierro, Margatina, Rust, and S{\o}gaard}]{hershcovich-etal-2022-challenges}
Daniel Hershcovich, Stella Frank, Heather Lent, Miryam de~Lhoneux, Mostafa Abdou, Stephanie Brandl, Emanuele Bugliarello, Laura Cabello~Piqueras, Ilias Chalkidis, Ruixiang Cui, Constanza Fierro, Katerina Margatina, Phillip Rust, and Anders S{\o}gaard. 2022.
\newblock \href {https://doi.org/10.18653/v1/2022.acl-long.482} {Challenges and strategies in cross-cultural {NLP}}.
\newblock In \emph{Proceedings of the 60th Annual Meeting of the Association for Computational Linguistics (Volume 1: Long Papers)}, pages 6997--7013, Dublin, Ireland. Association for Computational Linguistics.

\bibitem[{Hinrichs and Krauwer(2014)}]{hinrichs-krauwer-2014-clarin}
Erhard Hinrichs and Steven Krauwer. 2014.
\newblock \href {http://www.lrec-conf.org/proceedings/lrec2014/pdf/415_Paper.pdf} {The {CLARIN} research infrastructure: Resources and tools for e{H}umanities scholars}.
\newblock In \emph{Proceedings of the Ninth International Conference on Language Resources and Evaluation ({LREC}'14)}, pages 1525--1531, Reykjavik, Iceland. European Language Resources Association (ELRA).

\bibitem[{Hovy and Purschke(2018)}]{hovy-purschke-2018-capturing}
Dirk Hovy and Christoph Purschke. 2018.
\newblock \href {https://doi.org/10.18653/v1/D18-1469} {Capturing regional variation with distributed place representations and geographic retrofitting}.
\newblock In \emph{Proceedings of the 2018 Conference on Empirical Methods in Natural Language Processing}, pages 4383--4394, Brussels, Belgium. Association for Computational Linguistics.

\bibitem[{ISTAT(2017)}]{istat-2017-uso}
ISTAT. 2017.
\newblock L'uso della lingua italiana, dei dialetti e di altre lingue in {I}talia.
\newblock \url{https://www.istat.it/it/archivio/207961}.
\newblock Accessed: 2023-09-10.

\bibitem[{{Italian Law 482/1999}(1999)}]{law-482-1999}
{Italian Law 482/1999}. 1999.
\newblock Norme in materia di tutela delle minoranze linguistiche storiche.
\newblock \url{https://www.normattiva.it/uri-res/N2Ls?urn:nir:stato:legge:1999;482}.
\newblock Accessed: 2023-09-10.

\bibitem[{Jaber et~al.(2011)Jaber, Tonelli, and Delmonte}]{jaber-etal-2011-venetan}
Suhel Jaber, Sara Tonelli, and Rodolfo Delmonte. 2011.
\newblock \href {https://aclanthology.org/www.mt-archive.info/NLPSC-2011-Jaber.pdf} {Venetan to {E}nglish machine translation: Issues and possible solutions}.
\newblock In \emph{Proceedings of the 8th International NLPSC Workshop}, pages 69--80, Copenhagen, Denmark. Samfundslitteratur.

\bibitem[{Jaberg et~al.(1987)Jaberg, Jud, and Sanga}]{jaberg-jud-1987-ais}
Karl Jaberg, Jakob Jud, and Glauco Sanga. 1987.
\newblock \emph{Atlante Linguistico ed Etnografico dell'Italia e della Svizzera Meridionale}, {I}talian edition.
\newblock Unicopli, Milano, Italy.

\bibitem[{Joshi et~al.(2020)Joshi, Santy, Budhiraja, Bali, and Choudhury}]{joshi-etal-2020-state}
Pratik Joshi, Sebastin Santy, Amar Budhiraja, Kalika Bali, and Monojit Choudhury. 2020.
\newblock \href {https://doi.org/10.18653/v1/2020.acl-main.560} {The state and fate of linguistic diversity and inclusion in the {NLP} world}.
\newblock In \emph{Proceedings of the 58th Annual Meeting of the Association for Computational Linguistics}, pages 6282--6293, Online. Association for Computational Linguistics.

\bibitem[{Joulin et~al.(2017)Joulin, Grave, Bojanowski, and Mikolov}]{joulin-etal-2017-bag}
Armand Joulin, Edouard Grave, Piotr Bojanowski, and Tomas Mikolov. 2017.
\newblock \href {https://aclanthology.org/E17-2068} {Bag of tricks for efficient text classification}.
\newblock In \emph{Proceedings of the 15th Conference of the {E}uropean Chapter of the Association for Computational Linguistics: Volume 2, Short Papers}, pages 427--431, Valencia, Spain. Association for Computational Linguistics.

\bibitem[{King and Abney(2013)}]{king-abney-2013-labeling}
Ben King and Steven Abney. 2013.
\newblock \href {https://aclanthology.org/N13-1131} {Labeling the languages of words in mixed-language documents using weakly supervised methods}.
\newblock In \emph{Proceedings of the 2013 Conference of the North {A}merican Chapter of the Association for Computational Linguistics: Human Language Technologies}, pages 1110--1119, Atlanta, Georgia. Association for Computational Linguistics.

\bibitem[{Kreutzer et~al.(2022)Kreutzer, Caswell, Wang, Wahab, van Esch, Ulzii-Orshikh, Tapo, Subramani, Sokolov, Sikasote, Setyawan, Sarin, Samb, Sagot, Rivera, Rios, Papadimitriou, Osei, Suarez, Orife, Ogueji, Rubungo, Nguyen, M{\"u}ller, M{\"u}ller, Muhammad, Muhammad, Mnyakeni, Mirzakhalov, Matangira, Leong, Lawson, Kudugunta, Jernite, Jenny, Firat, Dossou, Dlamini, de~Silva, {\c{C}}abuk~Ball{\i}, Biderman, Battisti, Baruwa, Bapna, Baljekar, Azime, Awokoya, Ataman, Ahia, Ahia, Agrawal, and Adeyemi}]{kreutzer-etal-2022-quality}
Julia Kreutzer, Isaac Caswell, Lisa Wang, Ahsan Wahab, Daan van Esch, Nasanbayar Ulzii-Orshikh, Allahsera Tapo, Nishant Subramani, Artem Sokolov, Claytone Sikasote, Monang Setyawan, Supheakmungkol Sarin, Sokhar Samb, Beno{\^\i}t Sagot, Clara Rivera, Annette Rios, Isabel Papadimitriou, Salomey Osei, Pedro~Ortiz Suarez, Iroro Orife, Kelechi Ogueji, Andre~Niyongabo Rubungo, Toan~Q. Nguyen, Mathias M{\"u}ller, Andr{\'e} M{\"u}ller, Shamsuddeen~Hassan Muhammad, Nanda Muhammad, Ayanda Mnyakeni, Jamshidbek Mirzakhalov, Tapiwanashe Matangira, Colin Leong, Nze Lawson, Sneha Kudugunta, Yacine Jernite, Mathias Jenny, Orhan Firat, Bonaventure F.~P. Dossou, Sakhile Dlamini, Nisansa de~Silva, Sakine {\c{C}}abuk~Ball{\i}, Stella Biderman, Alessia Battisti, Ahmed Baruwa, Ankur Bapna, Pallavi Baljekar, Israel~Abebe Azime, Ayodele Awokoya, Duygu Ataman, Orevaoghene Ahia, Oghenefego Ahia, Sweta Agrawal, and Mofetoluwa Adeyemi. 2022.
\newblock \href {https://doi.org/10.1162/tacl_a_00447} {Quality at a glance: An audit of web-crawled multilingual datasets}.
\newblock \emph{Transactions of the Association for Computational Linguistics}, 10:50--72.

\bibitem[{Lekakou et~al.(2013)Lekakou, Baldiserra, and Anastasopoulos}]{lekakou-etal-2013-documentation}
Marika Lekakou, Valeria Baldiserra, and Antonis Anastasopoulos. 2013.
\newblock Documentation and analysis of an endangered language: Aspects of the grammar of {G}riko.
\newblock \url{http://griko.project.uoi.gr/}.
\newblock Accessed: 2023-05-01.

\bibitem[{Liu et~al.(2020)Liu, Gu, Goyal, Li, Edunov, Ghazvininejad, Lewis, and Zettlemoyer}]{liu-etal-2020-multilingual-denoising}
Yinhan Liu, Jiatao Gu, Naman Goyal, Xian Li, Sergey Edunov, Marjan Ghazvininejad, Mike Lewis, and Luke Zettlemoyer. 2020.
\newblock \href {https://doi.org/10.1162/tacl_a_00343} {Multilingual denoising pre-training for neural machine translation}.
\newblock \emph{Transactions of the Association for Computational Linguistics}, 8:726--742.

\bibitem[{Liu et~al.(2022)Liu, Richardson, Hatcher, and Prud{'}hommeaux}]{liu-etal-2022-always}
Zoey Liu, Crystal Richardson, Richard Hatcher, and Emily Prud{'}hommeaux. 2022.
\newblock \href {https://doi.org/10.18653/v1/2022.acl-long.272} {Not always about you: Prioritizing community needs when developing endangered language technology}.
\newblock In \emph{Proceedings of the 60th Annual Meeting of the Association for Computational Linguistics (Volume 1: Long Papers)}, pages 3933--3944, Dublin, Ireland. Association for Computational Linguistics.

\bibitem[{Lusito et~al.(2023)Lusito, Ferrante, and Maillard}]{lusito-etal-2023-text2}
Stefano Lusito, Edoardo Ferrante, and Jean Maillard. 2023.
\newblock \href {https://aclanthology.org/2023.computel-1.14} {Text normalization for low-resource languages: {T}he case of {L}igurian}.
\newblock In \emph{Proceedings of the Sixth Workshop on the Use of Computational Methods in the Study of Endangered Languages}, pages 98--103, Remote. Association for Computational Linguistics.

\bibitem[{Lusito and Maillard(2021)}]{lusito-maillard-2021-universal}
Stefano Lusito and Jean Maillard. 2021.
\newblock \href {https://aclanthology.org/2021.udw-1.10} {A {U}niversal {D}ependencies corpus for {L}igurian}.
\newblock In \emph{Proceedings of the Fifth Workshop on Universal Dependencies (UDW, SyntaxFest 2021)}, pages 121--128, Sofia, Bulgaria. Association for Computational Linguistics.

\bibitem[{Mager et~al.(2021)Mager, Oncevay, Rios, Ruiz, Palmer, Neubig, and Kann}]{americasnlp-2021-natural}
Manuel Mager, Arturo Oncevay, Annette Rios, Ivan Vladimir~Meza Ruiz, Alexis Palmer, Graham Neubig, and Katharina Kann, editors. 2021.
\newblock \href {https://aclanthology.org/2021.americasnlp-1.0} {\emph{Proceedings of the First Workshop on Natural Language Processing for Indigenous Languages of the Americas}}. Association for Computational Linguistics, Online.

\bibitem[{Maiden and Parry(1997)}]{maiden1997dialects}
Martin Maiden and Mair Parry. 1997.
\newblock \emph{The Dialects of Italy}.
\newblock Routledge, London, England.

\bibitem[{Malkin et~al.(2022)Malkin, Limisiewicz, and Stanovsky}]{malkin-etal-2022-balanced}
Dan Malkin, Tomasz Limisiewicz, and Gabriel Stanovsky. 2022.
\newblock \href {https://doi.org/10.18653/v1/2022.naacl-main.361} {A balanced data approach for evaluating cross-lingual transfer: Mapping the linguistic blood bank}.
\newblock In \emph{Proceedings of the 2022 Conference of the North American Chapter of the Association for Computational Linguistics: Human Language Technologies}, pages 4903--4915, Seattle, United States. Association for Computational Linguistics.

\bibitem[{de~Marneffe et~al.(2021)de~Marneffe, Manning, Nivre, and Zeman}]{de-marneffe-etal-2021-universal}
Marie-Catherine de~Marneffe, Christopher~D. Manning, Joakim Nivre, and Daniel Zeman. 2021.
\newblock \href {https://doi.org/10.1162/coli_a_00402} {{U}niversal {D}ependencies}.
\newblock \emph{Computational Linguistics}, 47(2):255--308.

\bibitem[{McCarthy et~al.(2019)McCarthy, Vylomova, Wu, Malaviya, Wolf-Sonkin, Nicolai, Kirov, Silfverberg, Mielke, Heinz, Cotterell, and Hulden}]{mccarthy-etal-2019-sigmorphon}
Arya~D. McCarthy, Ekaterina Vylomova, Shijie Wu, Chaitanya Malaviya, Lawrence Wolf-Sonkin, Garrett Nicolai, Christo Kirov, Miikka Silfverberg, Sabrina~J. Mielke, Jeffrey Heinz, Ryan Cotterell, and Mans Hulden. 2019.
\newblock \href {https://doi.org/10.18653/v1/W19-4226} {The {SIGMORPHON} 2019 shared task: Morphological analysis in context and cross-lingual transfer for inflection}.
\newblock In \emph{Proceedings of the 16th Workshop on Computational Research in Phonetics, Phonology, and Morphology}, pages 229--244, Florence, Italy. Association for Computational Linguistics.

\bibitem[{Miola(2013)}]{miola-2013-sociolinguistic}
Emanuele Miola. 2013.
\newblock \href {https://doi.org/10.1515/soci.2013.27.1.116} {A sociolinguistic account of {W}iki{P}iedmontese and {W}iki{L}ombard / {E}ine soziolinguistische auswertung von {W}iki-{P}iedmontesisch und {W}iki-{L}ombardisch / {E}tude sociolinguistique des {W}ikip{\'e}dias en {P}i{\'e}montais et en {L}ombard}.
\newblock \emph{Sociolinguistica}, 27(1):116--131.

\bibitem[{Miola(2017)}]{miola-2017-parola}
Emanuele Miola. 2017.
\newblock \href {http://hdl.handle.net/11585/669026} {Dalla parola alla scrittura: Il caso di {E}miliano, {V}eneto e {S}iciliano}.
\newblock \emph{Quaderni di Linguistica (La Scrittura all'Ombra della Parola)}, 5:59--72.

\bibitem[{Moseley(2010)}]{moseley2010atlas}
Christopher Moseley. 2010.
\newblock \href {https://unesdoc.unesco.org/ark:/48223/pf0000187026} {\emph{Atlas of the World's Languages in Danger}}, 3rd edition.
\newblock Memory of Peoples. UNESCO Publishing, Paris, France.

\bibitem[{Muresan et~al.(2022)Muresan, Nakov, and Villavicencio}]{acl-2022-association-linguistics-1}
Smaranda Muresan, Preslav Nakov, and Aline Villavicencio, editors. 2022.
\newblock \href {https://aclanthology.org/2022.acl-long.0} {\emph{Proceedings of the 60th Annual Meeting of the Association for Computational Linguistics (Volume 1: Long Papers)}}. Association for Computational Linguistics, Dublin, Ireland.

\bibitem[{Nguyen et~al.(2021)Nguyen, Rosseel, and Grieve}]{nguyen-etal-2021-learning2}
Dong Nguyen, Laura Rosseel, and Jack Grieve. 2021.
\newblock \href {https://doi.org/10.18653/v1/2021.naacl-main.50} {On learning and representing social meaning in {NLP}: {A} sociolinguistic perspective}.
\newblock In \emph{Proceedings of the 2021 Conference of the North American Chapter of the Association for Computational Linguistics: Human Language Technologies}, pages 603--612, Online. Association for Computational Linguistics.

\bibitem[{{NLLB Team} et~al.(2022){NLLB Team}, Costa-jussà, Cross, Çelebi, Elbayad, Heafield, Heffernan, Kalbassi, Lam, Licht, Maillard, Sun, Wang, Wenzek, Youngblood, Akula, Barrault, Gonzalez, Hansanti, Hoffman, Jarrett, Sadagopan, Rowe, Spruit, Tran, Andrews, Ayan, Bhosale, Edunov, Fan, Gao, Goswami, Guzmán, Koehn, Mourachko, Ropers, Saleem, Schwenk, and Wang}]{costa-etal-2022-no}
{NLLB Team}, Marta~R. Costa-jussà, James Cross, Onur Çelebi, Maha Elbayad, Kenneth Heafield, Kevin Heffernan, Elahe Kalbassi, Janice Lam, Daniel Licht, Jean Maillard, Anna Sun, Skyler Wang, Guillaume Wenzek, Al~Youngblood, Bapi Akula, Loic Barrault, Gabriel~Mejia Gonzalez, Prangthip Hansanti, John Hoffman, Semarley Jarrett, Kaushik~Ram Sadagopan, Dirk Rowe, Shannon Spruit, Chau Tran, Pierre Andrews, Necip~Fazil Ayan, Shruti Bhosale, Sergey Edunov, Angela Fan, Cynthia Gao, Vedanuj Goswami, Francisco Guzmán, Philipp Koehn, Alexandre Mourachko, Christophe Ropers, Safiyyah Saleem, Holger Schwenk, and Jeff Wang. 2022.
\newblock \href {https://doi.org/10.48550/arXiv.2207.04672} {{N}o {L}anguage {L}eft {B}ehind: {S}caling human-centered machine translation}.
\newblock \emph{CoRR}, arXiv:2207.04672v3.

\bibitem[{Ojha et~al.(2022)Ojha, Ahmadi, Liu, and McCrae}]{eurali-2022-resources}
Atul~Kr. Ojha, Sina Ahmadi, Chao-Hong Liu, and John~P. McCrae, editors. 2022.
\newblock \href {https://aclanthology.org/2022.eurali-1.0} {\emph{Proceedings of the Workshop on Resources and Technologies for Indigenous, Endangered and Lesser-resourced Languages in Eurasia within the 13th Language Resources and Evaluation Conference}}. European Language Resources Association, Marseille, France.

\bibitem[{Pellegrini(1977)}]{pellegrini-1977-carta}
Giovan~Battista Pellegrini. 1977.
\newblock \emph{{Carta dei Dialetti d'Italia}}.
\newblock {Profilo dei Dialetti Italiani}. {Pacini}, {Pisa, Italy}.

\bibitem[{Pellegrino(2021)}]{pellegrino-2021-greek}
Manuela Pellegrino. 2021.
\newblock \emph{Greek Language, Italian Landscape: Griko and the Re-storying of a Linguistic Minority}.
\newblock Hellenic Studies Series. Harvard University, Center for Hellenic Studies, Washington, DC, USA.

\bibitem[{Ramponi and Casula(2023)}]{ramponi-casula-2023-diatopit}
Alan Ramponi and Camilla Casula. 2023.
\newblock \href {https://aclanthology.org/2023.vardial-1.19} {{D}iatop{I}t: A corpus of social media posts for the study of diatopic language variation in {I}taly}.
\newblock In \emph{Tenth Workshop on NLP for Similar Languages, Varieties and Dialects (VarDial 2023)}, pages 187--199, Dubrovnik, Croatia. Association for Computational Linguistics.

\bibitem[{Rigoni~Stern(1998)}]{rigonistern-1998-sentieri}
Mario Rigoni~Stern. 1998.
\newblock \emph{Sentieri sotto la Neve}.
\newblock Supercoralli. Einaudi, Torino, Italy.

\bibitem[{Scherrer et~al.(2023)Scherrer, Jauhiainen, Ljube{\v{s}}i{\'c}, Nakov, Tiedemann, and Zampieri}]{vardial-2023-nlp}
Yves Scherrer, Tommi Jauhiainen, Nikola Ljube{\v{s}}i{\'c}, Preslav Nakov, J{\"o}rg Tiedemann, and Marcos Zampieri, editors. 2023.
\newblock \href {https://aclanthology.org/2023.vardial-1.0} {\emph{Tenth Workshop on NLP for Similar Languages, Varieties and Dialects (VarDial 2023)}}. Association for Computational Linguistics, Dubrovnik, Croatia.

\bibitem[{Schwartz(2022)}]{schwartz-2022-primum}
Lane Schwartz. 2022.
\newblock \href {https://doi.org/10.18653/v1/2022.acl-short.82} {{P}rimum {N}on {N}ocere: {B}efore working with {I}ndigenous data, the {ACL} must confront ongoing colonialism}.
\newblock In \emph{Proceedings of the 60th Annual Meeting of the Association for Computational Linguistics (Volume 2: Short Papers)}, pages 724--731, Dublin, Ireland. Association for Computational Linguistics.

\bibitem[{Schwenk et~al.(2021)Schwenk, Chaudhary, Sun, Gong, and Guzm{\'a}n}]{schwenk-etal-2021-wikimatrix}
Holger Schwenk, Vishrav Chaudhary, Shuo Sun, Hongyu Gong, and Francisco Guzm{\'a}n. 2021.
\newblock \href {https://doi.org/10.18653/v1/2021.eacl-main.115} {{W}iki{M}atrix: Mining 135{M} parallel sentences in 1620 language pairs from {W}ikipedia}.
\newblock In \emph{Proceedings of the 16th Conference of the European Chapter of the Association for Computational Linguistics: Main Volume}, pages 1351--1361, Online. Association for Computational Linguistics.

\bibitem[{Simons and Bird(2003)}]{simons-and-bird-2003-olac}
Gary Simons and Steven Bird. 2003.
\newblock \href {https://doi.org/10.1093/llc/18.2.117} {{The Open Language Archives Community: An infrastructure for distributed archiving of language resources}}.
\newblock \emph{Literary and Linguistic Computing}, 18(2):117--128.

\bibitem[{Thomason(2015)}]{thomason-2015-endangered}
Sarah~G. Thomason. 2015.
\newblock \href {https://doi.org/10.1017/CBO9781139033817} {\emph{Endangered Languages: An Introduction}}.
\newblock Cambridge Textbooks in Linguistics. Cambridge University Press, Cambridge, England.

\bibitem[{Tiedemann(2012)}]{tiedemann-2012-parallel}
J{\"o}rg Tiedemann. 2012.
\newblock \href {http://www.lrec-conf.org/proceedings/lrec2012/pdf/463_Paper.pdf} {Parallel data, tools and interfaces in {OPUS}}.
\newblock In \emph{Proceedings of the Eighth International Conference on Language Resources and Evaluation ({LREC}'12)}, pages 2214--2218, Istanbul, Turkey. European Language Resources Association (ELRA).

\bibitem[{Tonelli et~al.(2010)Tonelli, Pianta, Delmonte, and Brunelli}]{tonelli-etal-2010-venpro2}
Sara Tonelli, Emanuele Pianta, Rodolfo Delmonte, and Michele Brunelli. 2010.
\newblock \href {http://www.lrec-conf.org/proceedings/lrec2010/pdf/556_Paper.pdf} {{V}en{P}ro: {A} morphological analyzer for {V}enetan}.
\newblock In \emph{Proceedings of the Seventh International Conference on Language Resources and Evaluation ({LREC}'10)}, Valletta, Malta. European Language Resources Association (ELRA).

\bibitem[{Tyers et~al.(2017)Tyers, i~Font, Fronteddu, and Mart{\'\i}n-Mor}]{tyers-etal-2017-rule}
Francis~M Tyers, H{\`e}ctor~Al{\`o}s i~Font, Gianfranco Fronteddu, and Adri{\`a} Mart{\'\i}n-Mor. 2017.
\newblock \href {https://doi.org/10.1515/pralin-2017-0022} {Rule-based machine translation for the {I}talian-{S}ardinian language pair}.
\newblock \emph{The Prague Bulletin of Mathematical Linguistics}, 108:221--232.

\bibitem[{Vylomova et~al.(2020)Vylomova, White, Salesky, Mielke, Wu, Ponti, Maudslay, Zmigrod, Valvoda, Toldova, Tyers, Klyachko, Yegorov, Krizhanovsky, Czarnowska, Nikkarinen, Krizhanovsky, Pimentel, Torroba~Hennigen, Kirov, Nicolai, Williams, Anastasopoulos, Cruz, Chodroff, Cotterell, Silfverberg, and Hulden}]{vylomova-etal-2020-sigmorphon}
Ekaterina Vylomova, Jennifer White, Elizabeth Salesky, Sabrina~J. Mielke, Shijie Wu, Edoardo~Maria Ponti, Rowan~Hall Maudslay, Ran Zmigrod, Josef Valvoda, Svetlana Toldova, Francis Tyers, Elena Klyachko, Ilya Yegorov, Natalia Krizhanovsky, Paula Czarnowska, Irene Nikkarinen, Andrew Krizhanovsky, Tiago Pimentel, Lucas Torroba~Hennigen, Christo Kirov, Garrett Nicolai, Adina Williams, Antonios Anastasopoulos, Hilaria Cruz, Eleanor Chodroff, Ryan Cotterell, Miikka Silfverberg, and Mans Hulden. 2020.
\newblock \href {https://doi.org/10.18653/v1/2020.sigmorphon-1.1} {{SIGMORPHON} 2020 shared task 0: Typologically diverse morphological inflection}.
\newblock In \emph{Proceedings of the 17th SIGMORPHON Workshop on Computational Research in Phonetics, Phonology, and Morphology}, pages 1--39, Online. Association for Computational Linguistics.

\bibitem[{Wada et~al.(2021)Wada, Iwata, Matsumoto, Baldwin, and Lau}]{wada-etal-2021-learning}
Takashi Wada, Tomoharu Iwata, Yuji Matsumoto, Timothy Baldwin, and Jey~Han Lau. 2021.
\newblock \href {https://doi.org/10.18653/v1/2021.mrl-1.2} {Learning contextualised cross-lingual word embeddings and alignments for extremely low-resource languages using parallel corpora}.
\newblock In \emph{Proceedings of the 1st Workshop on Multilingual Representation Learning}, pages 16--31, Punta Cana, Dominican Republic. Association for Computational Linguistics.

\bibitem[{Wdowiak(2022)}]{wdowiak-2022-recipe}
Eryk Wdowiak. 2022.
\newblock \href {https://doi.org/10.1007/978-3-031-10464-0_50} {A recipe for low-resource {NMT}}.
\newblock In \emph{Intelligent Computing (SAI 2022)}, pages 739--746, Cham, Switzerland. Springer International Publishing.

\bibitem[{Wenzek et~al.(2020)Wenzek, Lachaux, Conneau, Chaudhary, Guzm{\'a}n, Joulin, and Grave}]{wenzek-etal-2020-ccnet}
Guillaume Wenzek, Marie-Anne Lachaux, Alexis Conneau, Vishrav Chaudhary, Francisco Guzm{\'a}n, Armand Joulin, and Edouard Grave. 2020.
\newblock \href {https://aclanthology.org/2020.lrec-1.494} {{CCN}et: Extracting high quality monolingual datasets from web crawl data}.
\newblock In \emph{Proceedings of the Twelfth Language Resources and Evaluation Conference}, pages 4003--4012, Marseille, France. European Language Resources Association.

\end{thebibliography}
